\documentclass[journal]{IEEEtran}

\usepackage[compress]{cite}
\usepackage{times}
\usepackage{epsfig}
\usepackage{amssymb}
\usepackage{url}
\usepackage{amsmath}
\usepackage{float}
\usepackage{graphicx}
\usepackage{amsfonts}
\usepackage{verbatim}
\usepackage{enumitem}
\usepackage{multirow}
\usepackage{dblfloatfix}
\usepackage{algorithm}
\usepackage{algpseudocode}
\usepackage{forloop}
\usepackage{subcaption}
\usepackage{color}

\newcommand{\bfx}{{\bf x}}

\newcommand{\bfd}{{\bf d}}
\newcommand{\bfs}{{\bf s}}
\newcommand{\bfy}{{\bf y}}
\newcommand{\bfe}{{\bf e}}
\newcommand{\reals}{\mathbb{R}}
\newcommand{\high}{\textrm{high}}
\newcommand{\low}{\textrm{low}}
\newcommand{\ord}{\mathcal{O}}

\begin{document}
\title{Cross-Scale Predictive Dictionaries}

\author{Vishwanath~Saragadam,~\IEEEmembership{Student Member,~IEEE,}
        ~Xin~Li,~\IEEEmembership{Fellow,~IEEE,}\\
        and~Aswin~C.\ Sankaranarayanan,~\IEEEmembership{Senior Member,~IEEE}% <-this % stops a space
\thanks{V.\ Saragaram and A.\ C.\ Sankaranarayanan are with the ECE Department at the Carnegie Mellon University.}%
\thanks{X.\ Li is with the ECE Department at the Duke Kunshan University.}}%
%\markboth{Transactions on Image Processing, ~Vol.~14, No.~8, December~2016}%

\maketitle

\begin{abstract}
Sparse representations using data dictionaries provide an efficient model particularly for signals that do not enjoy alternate analytic sparsifying transformations.
However, solving inverse problems with  sparsifying dictionaries can be computationally expensive, especially when the dictionary under consideration has a large number of atoms.
In this paper, we incorporate additional structure on to dictionary-based sparse representations for visual signals  to enable  speedups when solving sparse approximation problems.
The specific structure that we endow onto sparse models is that of a multi-scale modeling where the sparse representation at each scale is constrained by the sparse representation at coarser scales.
We show that this cross-scale predictive model delivers significant speedups, often in the range of 10-60$\times$, with little loss in accuracy for linear inverse problems associated with images, videos, and light fields.
\end{abstract}


\IEEEpeerreviewmaketitle

%----------- INTRODUCTION ------------ %
\section{Introduction}
% Introduction file.

Images are strongly correlated across scales, a fact that is often modeled and exploited to enhance image processing algorithms \cite{adelson1991pyramids,secker2003lifting}.
An important example of this idea is the wavelet tree model which provides a sparse as well as a \textit{predictive} model for the occurrence of non-zero wavelet coefficients across spatial scales \cite{wainwright2001random}.
The wavelet tree model arranges the wavelet coefficients of an image onto a tree whose nodes  correspond to the coefficients and  each level corresponds to coefficients associated with a particular scale.
Under such an organization, the dominant non-zero  coefficients form a connected rooted sub-tree \cite{baraniuk1999optimal}, i.e.,
children of a node with  small wavelet coefficients are expected to take small values as well.
This property has found widespread applicability in tasks like compression \cite{shapiro1993embedded}, sensing \cite{baraniuk2010model,deutsch2009adaptive}, and processing  \cite{baraniuk1999optimal}.
While the wavelet tree model provides excellent approximation capabilities for images, similar models with cross-scale predictive property are largely unknown for other visual signals including videos, hyperspectral images, and light fields.

Overcomplete dictionaries provide an alternate approach for enabling sparse representations \cite{olshausen1997sparse}.
Given a large amount of  data, we can  \textit{learn} a dictionary such that the training dataset can be expressed as a sparse linear combination of the elements/atoms of the dictionary.
The reliance on learning, as opposed to analytic constructions as in the case of wavelets, provides immense flexibility towards obtaining a dictionary that is tuned to the specifics of a particular signal class.
Overcomplete dictionaries have found wide applicability for sensing and processing of images \cite{aharon2006img}, videos \cite{hitomi2011video}, light fields \cite{marwah2012compressive}, and other visual signals \cite{hui2015dictionary}.
However, not much attention has been paid to the incorporation of predictive models to enable speed ups by exploiting correlations across spatial, temporal and angular scales. 
\begin{figure}[!t]
\centering
\includegraphics[width=0.155\textwidth]{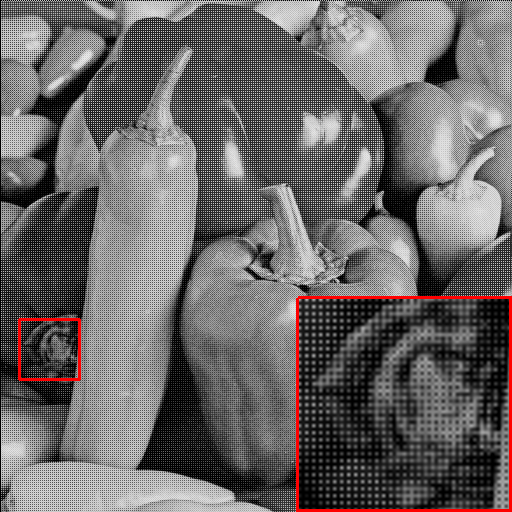}
\includegraphics[width=0.155\textwidth]{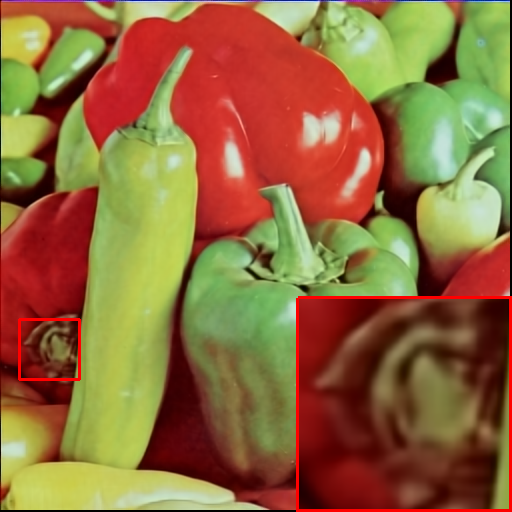}
\includegraphics[width=0.155\textwidth]{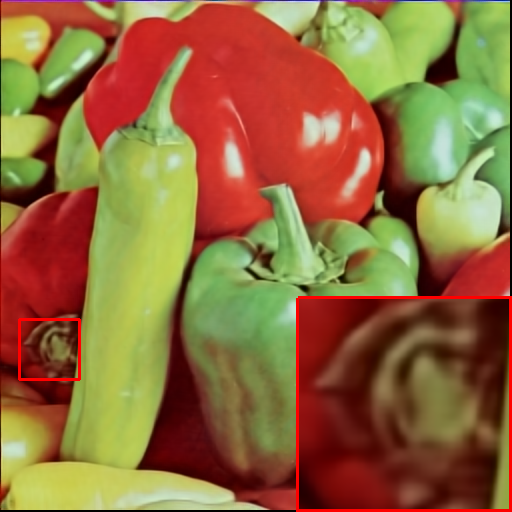}
\caption{Left to right: Bayer image, image reconstructed using OMP, and image reconstructed using the proposed method. While OMP takes $16$ minutes, the proposed method takes only $1.5$ minutes with little loss in reconstruction quality.}
\label{fig:bayer}
\end{figure}

In this paper, we propose incorporation of cross-scale predictive modeling in sparsifying dictionaries, thus combining class-specific adaptation with speed ups offered by predictive models.
Our specific contributions are as follows.
\begin{itemize}[leftmargin=*]
	\item \textbf{Model.} We propose a novel signal model that uses multi-scale sparsifying dictionaries  to provide cross-scale prediction for a wide array of visual signals. Specifically, given the set of sparsifying dictionaries --- one for each scale --- the non-zero support patterns of a signal  and its downsampled counterparts are constrained to only exhibit specific pre-determined patterns.
	\item \textbf{Computational speedups.} The proposed signal model, with its constrained support pattern across scales, naturally enables cross-scale prediction that can be used to speedup the runtime of algorithms like orthogonal matching pursuit (OMP) \cite{pati1993orthogonal}.  Figure \ref{fig:bayer} shows speed-ups obtained for demosaicing of images; here, we obtain a $10\times$ speed up with little  loss in accuracy over a similar-sized dictionary. 
	\item \textbf{Learning.} Given large collections of training data, we propose a simple training method, which is modified from the classical K-SVD algorithm \cite{aharon2006img}, to obtain dictionaries that are consistent with our proposed model.
	\item \textbf{Validation.} We verify empirically that the model works through simulation on an array of visual signals including images, videos, and light field images. 
\end{itemize}

A shorter version of this paper appeared at the IEEE International Conference on Image Processing \cite{saragadam2016cross}. This journal paper extends the results to a larger class of signals including light fields as well as shows results on real data captured from compressive imaging hardware.

%----------- PRIOR WORK ------------ %
\section{Prior work} \label{sec:prior}
\subsection{Notation} We denote vectors in bold font and scalars/matrices in capital letters. A vector is said to be $K$-sparse if it has at most $K$ non-zero entires. 
The support of a sparse vector $\bfs$, denoted as  $\Omega_\bfs$, is the set of the indices of its non-zero entries.
The $\ell_0$-norm of a sparse vector is the number of non-zero entries or equivalently the cardinality of its support.
Finally, given a dictionary $D \in \reals^{N \times T}$ and a support set $\Omega$, $D_{|\Omega}$ refers to the matrix of size $N\times | \Omega |$ formed by selecting columns of $D$ corresponding to the elements of $\Omega$; similarly, given a vector $\bfs$, $\bfs_{|\Omega}$ refers to an $|\Omega|$-dimensional vector formed by selecting entries in $\bfs$ corresponding to $\Omega$.

\subsection{Sparse approximation}
Sparse approximation problems arise in a wide range of settings \cite{elad2010sparse}.
The broad problem definition is as follows: given a vector $\bfx \in \reals^N$, a matrix $D \in \reals^{N \times T}$, we solve
\[ \textrm{(P0)} \qquad\qquad \min_{\bfs \in \reals^T} \| \bfx - D \bfs \|_2 \quad\textrm{s.t.} \quad \| \bfs \|_0 \le K.\]
While the problem itself is NP-hard \cite{natarajan1995sparse}, there are many greedy and relaxed approaches to solving $\textrm{(P0)}$.
Of particular interest to this paper is OMP \cite{pati1993orthogonal}, a greedy approach  to solving $\textrm{(P0)}$.
OMP recovers the support of the sparse vector $\bfs$, one element at a time, by finding the column of the dictionary that is most correlated with the current residue.
In each iteration of the algorithm, there are three steps: first, the index of the atom that is closest in angle to the current residue is added to the support; second, solving a least square problem with the updated support to obtain the current estimate; and third, updating the residue by removing the contribution of the current estimate. Outline of the procedure is given in Algorithm \ref{alg:omp}.

% OMP algorithm

\begin{algorithm}[h]
	\caption{Orthogonal matching pursuit}
	\label{alg:omp}
	\begin{algorithmic}
		\Require $\bfx$, $D$, $K$
		\State $r \leftarrow \bfx$
		\State $\Omega \leftarrow \phi$
		\State $\alpha \leftarrow \phi$
		
		\For{$n = 1$ to $K$}
			\State $k \leftarrow \arg \max_{i} |\langle\bfd_i, r\rangle|$ \Comment{(Proxy)}
			\State $\Omega \leftarrow \Omega \bigcup k$ \Comment{(Support merge)}
			\State $\alpha \leftarrow \arg \min_{\beta} \| \bfx - D_{|\Omega} \beta \|^2$ \Comment{(Projection)}
			\State $r \leftarrow \bfx - D_{|\Omega} \alpha$ \Comment{(Residue)}
		\EndFor
		
		\State \Return $\Omega$, $\alpha$
	\end{algorithmic}
\end{algorithm}

The proxy step and the projection step are the two computationally intensive steps in OMP. The time complexity of the proxy step is $\mathcal{O}(NTK)$, while that of the projection step is $\mathcal{O}(NK^3)$ for $K$ iterations. For very large dictionaries and very sparse representation, the proxy step is the dominating term, which grows linearly with dictionary size.
\begin{figure}[!ttt]
	\centering
	\includegraphics[width=0.4\textwidth]{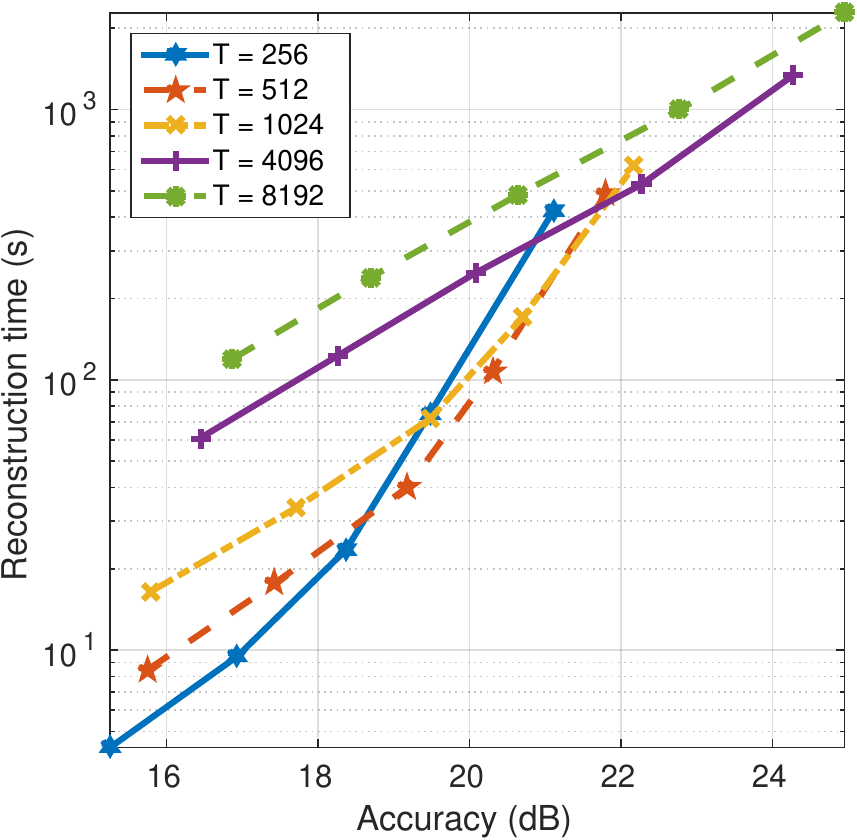}
	\caption{Time versus  accuracy for varying dictionary size when denoising $8\times8\times32$ video patches. Each curve was generated by varying the sparsity level, $K$, from 8 to 64 in multiples of 2. We observe that it is computationally beneficial to use a dictionary with a larger number of atoms at a smaller sparsity level as opposed to a smaller dictionary at a higher sparsity.}
	\label{fig:nsweep}
\end{figure}
\subsection{Speeding up OMP} 
A number of techniques have been devoted to speed up different steps of OMP.
For problems in high-dimensions, i.e.\ large values of $N$, one approach is to project to a lower dimension by obtaining random projections of the dictionary \cite{vitaladevuni2011efficient}.
Specifically, as opposed to the objective $\| \bfx - D \bfs \|_2$, we minimize $\| \Phi \bfx - \Phi D \bfs \|_2$ where $\Phi \in \reals^{M\times N}$, $M < N$, is a random matrix that preserves the geometry of the problem thereby allowing us to perform all computations in an $M$-dimensional space.
In the context of high-dimensional data, it is typical to have dictionaries with very large number of atoms, i.e $T \gg N$, and in such a setting, the proxy step becomes a bottle neck.
The need for large dictionaries is driven by a requirement for higher accuracy of reconstruction. Empirically, larger dictionaries are needed for higher accuracy, which is evident from the time vs accuracy plot in Figure \ref{fig:nsweep} for denoising of videos.

\subsection{Multi-scale approaches for sparse approximation}
Various methods have been proposed to speed up sparse approximation by imposing structure on the coefficients. Such methods employ a tree-like arrangement of the sparse coefficients which give it a logarithmic complexity improvement.

One approach is by using approximate nearest neighbors and shallow-tree based matching to speed up the proxy step  \cite{ayremlou2014fast, gribonval2001fast}.
Instead of searching across all elements of the dictionary, the dictionary is arranged into a shallow tree for fast search. In certain conditions, an $\mathcal{O}(\log(T))$ search complexity can be obtained.
However, this results in a reduction in accuracy, as the closest dictionary atom is computed through approximate nearest neighbor method.

Another approach is to restrict the search space by imposing a tree structure on sparse coefficients \cite{la2006tree}. Complexity of the proxy step would then reduce to $\ord{(\log(T))}$.
Restricting search space through prediction has been explored in \cite{gribonval2001fast} for approximation by chirplet atoms.
Here, the method first finds an approximation to the input signal through a gabor atom and, subsequently, the scale and chirp parameters are optimized locally. %to get an approximate chirplet atom.
Though such methods provide significant speedups, their usage is restricted to signals with known structure, such as wavelets for images or hierarchical structure of chirplet atoms for sound.

\subsection{Dictionary learning}
For signal classes that have no obvious sparsifying transforms, a promising approach is to \textit{learn} a dictionary that provides sparse representations for the specific signal class of interest.
Field and Olshausen \cite{olshausen1997sparse}, in a seminal contribution, showed that patches of natural images were sparsified by a dictionary containing Gabor-like atoms --- this provided a connection between sparse coding and the receptor fields in the visual cortex.
More recently,  Aharon et al.\ \cite{aharon2006img} proposed the ``K-SVD'' algorithm which can be viewed as an extension of the k-means clustering algorithm for dictionary learning.
Given a collection of training data $X = [ \bfx_1, \bfx_2, \ldots, \bfx_Q ], \bfx_i \in \reals^N$, K-SVD aims to learn a dictionary $D \in \reals^{N \times T}$ such that $X \approx D [ \bfs_1, \ldots, \bfs_Q]$ with each $\bfs_k$ being $K$-sparse.
This was one of the first forays into learning good dictionaries for sparse representation. However, with increasing complexity and signal dimension, larger dictionaries are needed, which requires larger computation time.
%
%We showed empirically in Figure \ref{fig:nsweep} that larger dictionaries are required for increasing dimension of signal, which can be a computationally expensive process.

\subsection{Multi-scale dictionary models}
Learning dictionary atoms that are innately clustered is an intuitive way of speeding up the approximation process with large dictionaries. Particularly for visual signals, clustering by incorporating scale or spatial complexity of the signal has been explored before.
Jayaraman et al.\ \cite{thiagarajan2013learning} learn dictionaries by a multi-level representation of image patches where simple patches are captured in the early stages while more complex textures are only resolved at the higher levels.
This provides speedups when solving sparse approximation problems since patches that occur more often are captured at the earlier levels.
While speedups are constant when compared to a dictionary of the same size, it does not scale up well to high dimensional signals.
Similar to this work, we propose multiple levels of representation across scales that captures complex patterns at finer scales while also incorporating a predictive framework.

Imposing a tree structure on sparse coefficients to learn dictionaries has been explored in the context of images.
Jenatthon et al.\ \cite{jenatton2010proximal} present a hierarchical dictionary learning mechanism, where they impose a tree structure on the sparsity, which forces the dictionary atoms to cluster like a tree.
Though it does give higher accuracy of reconstruction, not much has been said about the speed up obtained.
Mairal et al.\ \cite{mairal2007multiscale}  learn a  dictionary based on quad-tree models, where each patch is further sub-divided into four non-overlapping patches.
While this method gives better accuracy, the algorithm is very slow, as it involves approximations of successive decomposition of a big image patch into smaller image patches.
None of the multi-scale learning algorithms exploit the cross-scale structure underlying  visual signals.

\subsection{Compressive sensing (CS)} An application of sparse representations is in CS where signals are sensed from far-fewer measurements than their dimensionality \cite{baraniuk2007compressive}.
CS relies on  low-dimensional representations for the sensed signal such as sparsity  under a transform or a dictionary.
There is a rich body of work on applying CS  to imaging of visual signals including images\cite{duarte2008single,chen2015fpa} videos \cite{hitomi2011video,reddy2011p2c2,sankaranarayanan2012cs} and light fields \cite{marwah2012compressive,tambe2013towards}.
Most relevant to our paper is the video CS work of Hitomi et al.\ \cite{hitomi2011video} where a sparsifying dictionary is used on video patches to recover high-speed videos from low-frame rate sensors. 
Hitomi et al.\ also demonstrated the accuracy enabled by very large dictionaries; specifically, they obtained remarkable results with a dictionary of  $T = 100,000$ atoms for video patches of dimension $N = 7 \times 7 \times 36 = 1764$. 
However, it is reported that  the recovery of $36$ frames of videos took more than an hour with a $100,000$ atom dictionary.
Clearly, there is a need for faster recovery techniques.

\subsection{Wavelet-tree model} Our proposed method is inspired by multi-resolution representations and tree-models enabled by wavelets.
Baraniuk \cite{baraniuk1999optimal} showed that the non-zero wavelet coefficients form a rooted sub-tree for signals that have trends (smooth variations) and anomalies (edges and discontinuities). 
Hence, piecewise-smooth signals enjoy a sparse representation with a structured support pattern with the non-zero wavelet coefficients  forming a rooted sub-tree. 
Similar properties have also been shown for 2D images under the separable Haar basis \cite{shapiro1993embedded}.
However, in spite of these elegant results for images, there are no obvious sparsifying bases for higher-dimensional visual signals like videos and light field images. 
To address this, we build cross-scale predictive models, similar to the wavelet tree model,  by replacing a basis with an over-complete dictionary that is capable of  providing a sparse and predictive representation for a wide class of signals.
%

%----------- PROPOSED SIGNAL MODEL ------------ %
\section{Cross-Scale Predictive Models} \label{sec:model}

\subsection{Proposed model}
\begin{figure}[!ttt]
	\centering
	\includegraphics[width=0.47\textwidth]{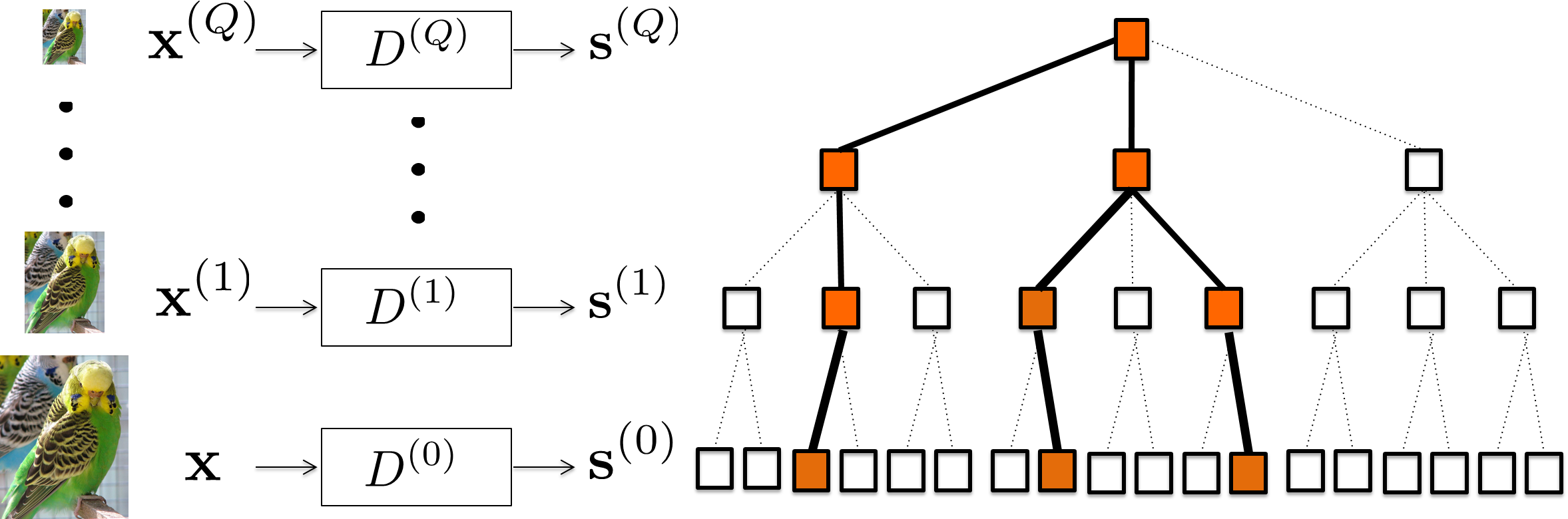}
	\caption{Proposed cross-scale signal model with sparse coefficients across scales forming a rooted subtree. We analyze the signal $\bfx$ at multiple scales such that $\bfx^{(k)}$ is obtained by downsampling it successively $k$-times. At the $k$-th scale, we learn a dictionary $D^{(k)}$ such that sparsifies the downsamped signal $\bfx^{(k)}$, i.e, $\bfx^{(k)} = D^{(k)} \bfs^{(k)}$. We arrange the sparse coefficients $\{  \bfs^{(k)} \}$  onto a tree and enforce the cross-scale prediction property as follows: a child atom can take non-zero values only if its parent is non-zero.}
	\label{fig:zerotree}
\end{figure}

The proposed signal model extends the notion of multi-resolution representation of signals beyond images. %Instead of relying on analytical bases for sparseness and prediction, we propose a signal model based on learned dictionaries which retain the properties of both overcomplete dictionaries and wavelet trees.
Given a signal $f$, we can represent it in the multi-resolution framework \cite{mallat1989theory}  as:
\begin{align*}
A^{2^j}f(x) &= \sum_{k=1}^{N_{2^j}}\lambda_k \phi^{2^j}_k(x)
			 = \Phi^{2^j} \Lambda^{2^j},
\end{align*}
where $A^{2^j}$ is the projection operator to the $2^j$ scale space, and $\{\phi^{2^j}_k(x)\}$ form a wavelet basis at $2^j$ scale.
For piecewise constant signals like images, the wavelet coefficients form a rooted sub-tree.
While it is hard to find such analytical bases for an arbitrary signal class, we can instead retain the multi-resolution  framework but replace the bases $\Phi^{2^j}$ with overcomplete dictionaries. 
Hence, we propose a signal model that predicts the support of a signal across scales (see Figure \ref{fig:zerotree}). We present our model with two-scale scenario for ease of understanding.
Given a collection of signals, $\mathcal{X} \subset \reals^N$, our proposed signal model consists of two sparsifying dictionaries, $D_\high \in \reals^{N \times T_\high}$ and  $D_\low \in \reals^{N_\low \times T_\low}$, that satisfy the following three properties.

\begin{itemize}[leftmargin=*]
\item \textit{Sparse approximation at the finer scale.} A signal $\bfx \in \mathcal{X}$ enjoys a $K_\high$-sparse representation in $D_\high$, i.e,  $\bfx \approx D_\high \bfs_\high$ with $\| \bfs_\high \|_0 \le K_\high$.
\item  \textit{Sparse approximation at the coarser scale.} Given $\bfx \in \mathcal{X}$ and a downsampling operator $W: \reals^N \mapsto \reals^{N_\low}$, the downsampled signal $\bfx_\low = W \bfx$ enjoys a sparse representation in $D_\low$, i.e., $\bfx_\low \approx D_\low \bfs_\low$ with $\| \bfs_\low \|_0 \le K_\low$. The downsampling operator $W$ is domain specific.
\item  \textit{Cross-scale prediction.} The support of $\bfs_\high$ is constrained by the support of $\bfs_\low$; specifically, $\Omega_{\bfs_\high} \subset f(\Omega_{\bfs_\low})$, where the mapping $f(\cdot)$ is known a priori.
\end{itemize}

We make a few observations.

\textit{Observation 1.} $T_\high \gg T_\low$ since $N_\high \gg N_\low$. With the increase of dimension of the signal, more complex patterns emerge which require larger number of redundant elements. Empirically we found that the number of atoms in a dictionary increases super linearly with increasing dimension of the signal for a given approximation accuracy (see Figure \ref{fig:nsweep}).

\textit{Observation 2.} Recall that the computational time of OMP is proportional to the number of atoms in the dictionary since, at each iteration of the algorithm, we need to compute the inner product between the residue and the atoms in the dictionary. If we can constrain the search space by constraining the number of atoms, then we can obtain computational speedups. 

The proposed model obtains speedups by first solving a sparse approximation problem  at the coarse scale and subsequently exploiting the cross-scale prediction property to constrain the support at the finer scale.
The source of the speedups relies on two intuitive ideas: first, solving a sparse approximation problem for a problem with fewer atoms (and in a smaller dimension) is faster due to OMP's runtime being linear in the number of atoms of the dictionary used\cite{mailhe2009low}; and second, if we knew the support of $\bfs_\low$, then we can simply discard all atoms in $D_\high$ that do not belong to $f(\Omega_{\bfs_\low})$ since the support of $\bfs_\high$ is guaranteed to lie within $f(\Omega_{\bfs_\low})$.

\subsection{Cross-scale mapping} We propose the following strategy for the cross-scale mapping $f$. 
Let $Q = T_\high/T_\low$ (assuming $T_\high$ and $T_\low$ are chosen to ensure $Q$ is an integer).
The cross-scale prediction map is defined using this simple rule.
\[  i \in \Omega_{\bfs_\low} \implies (i-1)Q+\left\{ 1, 2, \ldots, Q \right\} \subset f(\Omega_{\bfs_\low})\]
Each element of the support $\Omega_{\bfs_\low}$ in the coarser scale controls the inclusion/exclusion of a \textit{non-overlapping block} of locations for the sparse vector in the finer scale.
As a consequence, the cardinality of $f(\Omega_{\bfs_\low})$ is $Q K_\low$.
\subsection{Solving inverse problems under the proposed signal model.} We now detail the procedure for solving a sparse approximation problem using the proposed signal model (see Figure \ref{fig:zerotree}).
Specifically, we seek to recover $\bfx \in \mathcal{X}$ from a set of linear measurements $\bfy \in \reals^M$ of the form 
\[ \bfy =  \Phi \bfx  + \bfe = \Phi D_\high \bfs_\high + \bfe, \]
where $\Phi \in \reals^{M \times N}$ is the measurement matrix and $\bfe$ is the measurement noise.
As indicated earlier, we obtain $\bfs_\high$ using a two-step procedure.

\textit{Step1 --- Sparse approximation at the coarse scale.} We first solve the following sparse approximation problem:
\begin{align*} (P_\low) \quad\quad \widehat{\bfs}_\low = \arg\min_{\bfs_\low}  \| \bfy - \Phi U D_\low \bfs_\low \|_2 
\\
\textrm{s.t.} \quad\quad \| \bfs_\low \|_0 \le K_\low.
\end{align*}
Here, $U:\reals^M \mapsto \reals^N$ is an up-sampling operator such that $WU$ is an identity map on $\reals^{N_\low}$. In all our experiments, we used a uniform downsampler and a nearest-neighbour up sampler specific to the domain of the signal.
This step recovers a low-resolution approximation  to the signal, $\bfx_\low = D_\low \widehat{\bfs}_\low$.

\textit{Step 2 --- Sparse approximation at the finer scale.} Armed with the support  $\widehat{\Omega} =  \Omega_{\widehat{\bfs}_\low}$, we can solve for $\bfs_\high$ by solving:
\begin{align*}
  (P_\high) \quad\quad (\widehat{\bfs}_\high)_{|f(\Omega)} = \arg\min_{\alpha}  \| \bfy - \Phi (D_\high)_{|\Omega} \alpha \|_2 \\
  \textrm{s.t.} \quad\quad \| \alpha \|_0 \le K_\high. 
\end{align*}
The sparse approximation problems in both steps are solved using OMP. The proposed mapping across scales for the sparse support forms a zero tree, where a coefficient  is zero if the corresponding coefficient at coarser scale is zero. Hence we refer to our algorithm as zero tree OMP. Algorithm \ref{alg:zt_omp} outlines the zero tree OMP procedure.

% Zero tree OMP algorithm

\begin{algorithm}[h]
	\caption{Zero tree OMP}
	\label{alg:zt_omp}
	\begin{algorithmic}
		\Require $\bfx$, $D_\low$, $D_\high$, $K_\low$, $K_\high$, $W$
		
		\State $\bfx_\low \leftarrow W\bfx$
		\State $r_\low \leftarrow \bfx_\low$
		\State $\Omega_\low \leftarrow \phi$
		\State $\alpha_\low \leftarrow \phi$
		
		\For{$n = 1$ to $K_\low$}
			\State $k \leftarrow \arg \max_{i} |\langle(\bfd_\low)_i, r_\low\rangle|$
			\State $\Omega_\low \leftarrow \Omega_\low \bigcup k$
			\State $\alpha_\low \leftarrow \arg \min_{\beta} \| \bfx_\low - (D_\low)_{|\Omega} \beta \|^2$
			\State $r_\low \leftarrow \bfx_\low - (D_\low)_{|\Omega} \alpha_\low$
		\EndFor
		
		\State $\hat{D} = (D_\high)_{|f(\Omega_\low)}$
		
		\State $r_\high \leftarrow \bfx_\high$
		\State $\Omega_\high \leftarrow \phi$
		\State $\alpha_\high \leftarrow \phi$
		
		\For{$n = 1$ to $K_\high$}
		\State $k \leftarrow \arg \max_{i} |\langle\hat{\bfd}_i, r_\high\rangle|$
		\State $\Omega_\high \leftarrow \Omega_\high \bigcup k$
		\State $\alpha_\high \leftarrow \arg \min_{\beta} \| \bfx_\high - \hat{D}_{|\Omega} \beta \|^2$
		\State $r_\high \leftarrow \bfx_\high - \hat{D}_{|\Omega} \alpha_\high$
		\EndFor
		
		\State \Return $\Omega_\high$, $\alpha_\high$
	\end{algorithmic}
\end{algorithm}

\subsection{Theoretical speedup.} We provide expressions for the expected speedups over traditional single-scale OMP.
Since any analysis of speedup has to account for the complexity of implementing $\Phi$, we consider the denoising problem where $\Phi$ is the identity matrix.

Let $C(N, T, K)$ be the amount of time required to solve a sparse-approximation problem using OMP for a dictionary of size $N \times T$ and sparsity level $K$. 
Hence, obtaining $\bfs_\high$ directly from $\bfx$ would require $C(N, T_\high, K_\high)$ computations.
In contrast, our proposed two-step solution using cross-scale prediction has a computational cost of $C(N_\high, T_\low, K_\low)$ $+ C(N, Q K_\low, K_\high)$.

To compute the dependence of $C(N, T, K)$ on $N, T,$ and $K$, recall that for each iteration in the OMP algorithm, we need $\ord(NT)$ operations \cite{mailhe2009low} for finding inner product between the residue and the dictionary atoms, $\ord(T)$ operations to find the maximally aligned vector and $\ord(K^3 + K^2 N)$ operations for the least-squares step.  Thus,
\begin{align}
\label{eq:time_omp}
C(N, T, K) = \ord(NTK+TK+K^4 + K^3N). \nonumber
\end{align}
For dictionaries with a large number of atoms, i.e., large $T$, and small values for sparsity level $K$, the linear dependence on $T$ dominates the total computation time.
Hence, the speedup provided by our algorithm is approximately $T_\high/(T_\low + K_\low Q)$.
\subsection{Learning cross-scale sparse models.} We learn the dictionaries $(D_\high, D_\low)$ with a simple modification to the K-SVD algorithm.

\textit{Inputs.} The inputs to the learning/training phases are the training dataset $X = [ \bfx_1, \bfx_2, \ldots \bfx_n]$ and the values for the parameters $K_\high, K_\low, T_\high$, and $T_\low$.

\textit{Step 1 --- Learning $D_\low$.} We learn the coarse-scale dictionary $D_\low$ by  applying K-SVD to downsampled training dataset $X_\low = [ W \bfx_1, W\bfx_2, \ldots W\bfx_n]$. A by-product of learning the dictionary $D_\low$ are the supports $\{ \Omega_{s_\low, k_\low} \}$ of sparse approximations of the downsampled training dataset.

\textit{Step 2 --- Learning $D_\high$.} We learn the fine-scale dictionary $D_\high = [ \bfd_1, \ldots, \bfd_{T_\high} ]$ by solving
\begin{equation*}
\begin{aligned}
\label{eq:learnhigh}
 \min_{D_\high, S_\high}  \| Y - D_\high S_\high \|_F 
 \textrm{ s.t.}\quad\quad
 \| \bfd_k \|_2 = 1,\\ 
\textrm{support}(\bfs_\high) \subset f(\Omega_{\low, k})
\end{aligned}
\end{equation*}
The above optimization problem can be solved simply by modifying the sparse approximation step of K-SVD to restrict the support appropriately. 

As a consequence of speed up in approximation step, dictionary learning by proposed method is also faster. Recall that K-SVD alternates between dictionary learning and sparse approximation. Since the modified K-SVD algorithm replaces OMP by zero-tree OMP, the overall time taken for each iteration reduces, thus speeding up the learning of dictionaries.
\begin{figure}[!tt]
	\centering
	\begin{subfigure}[t]{0.45\columnwidth}
		\centering
		\includegraphics[width=\textwidth]{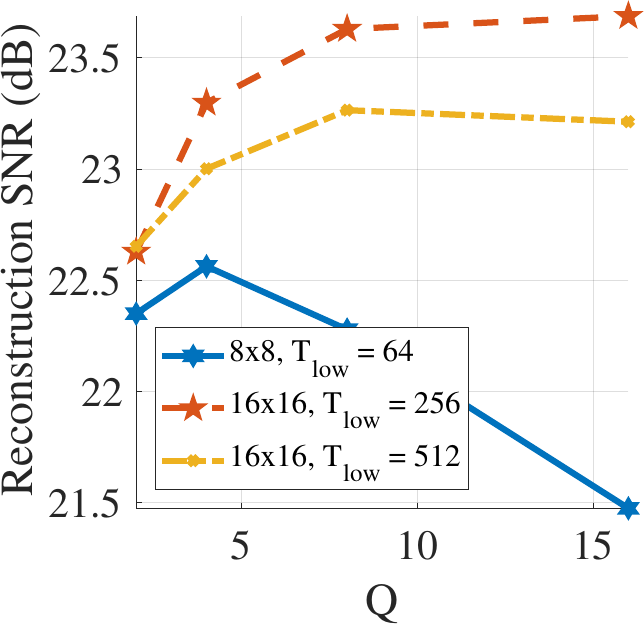}
		\caption{Image denoising}
	\end{subfigure}
	\quad
	\begin{subfigure}[t]{0.45\columnwidth}
		\centering
		\includegraphics[width=\textwidth]{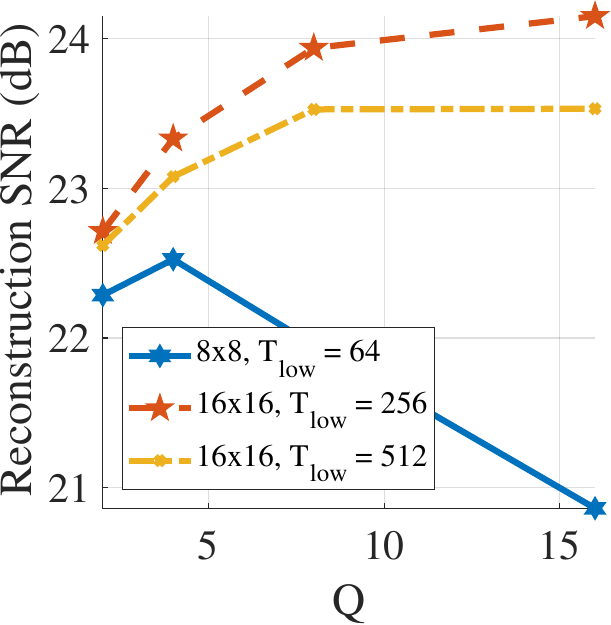}
		\caption{Image inpainting}
	\end{subfigure}
	\caption{Effect of $Q = {T_\high}/{T_\low}$ on approximation accuracy for various signal dimensions and dictionary sizes. For $8\times8$ patch, the optimal $Q$ is 4, while that for $16\times16$ is 16 for $T_\low = 256$ and 8 for $T_\low = 512$. Between the two dictionary sizes, $T_\low = 256$ gives better performance for $16\times16$ patches. Metrics such as these may be used for finding the optimal $Q$ and $T_\low$.}
	\label{fig:q_effect}
\end{figure}
\subsection{Parameter selection} The design parameters in the two scale dictionary training are $K_\low, K_\high, T_\low,$ and $T_\high$. $K_\low$ can be chosen to fine tune the accuracy at lower scales. For compressive sensing purposes, lower sparsity promises better reconstruction results. Hence a small $K_\low$ gives better results. We found that $K_\low$ in range of $2 - 4$ worked well. $K_\high$ should be greater than or equal to $K_\low$, as at least one atom corresponding to the low resolution atom will be picked.

The parameter $Q = \frac{T_\high}{T_\low}$ can be chosen by cross-validation. 
As an example, cross-validation may be performed with denoising as a test metric.
The value of $Q$ that gives highest reconstruction accuracy can then be chosen as the optimal $Q$.
To illustrate this, we trained dictionaries for various values of $T_\low$ and $Q$ for image patches of various sizes, and tested them for denoising and inpainting for the ``peppers" image.
Figure \ref{fig:q_effect} shows a comparison of approximation accuracy for various signal dimensions with different dictionary values as a function of $Q$.
For $8\times 8$ image patches, $Q = 4$ gives best results, whereas it is 16 for $16\times 16$ patches and 8 for $24 \times 24$ patches.
Since retraining dictionaries for each value of $Q$ is a time consuming process, $T_\low$ and $T_\high$ were chosen as would be appropriate for the signal dimension, $N_\low$ and $N_\high$ respectively.

\begin{figure*}
	\centering
	\includegraphics[width=0.95\textwidth]{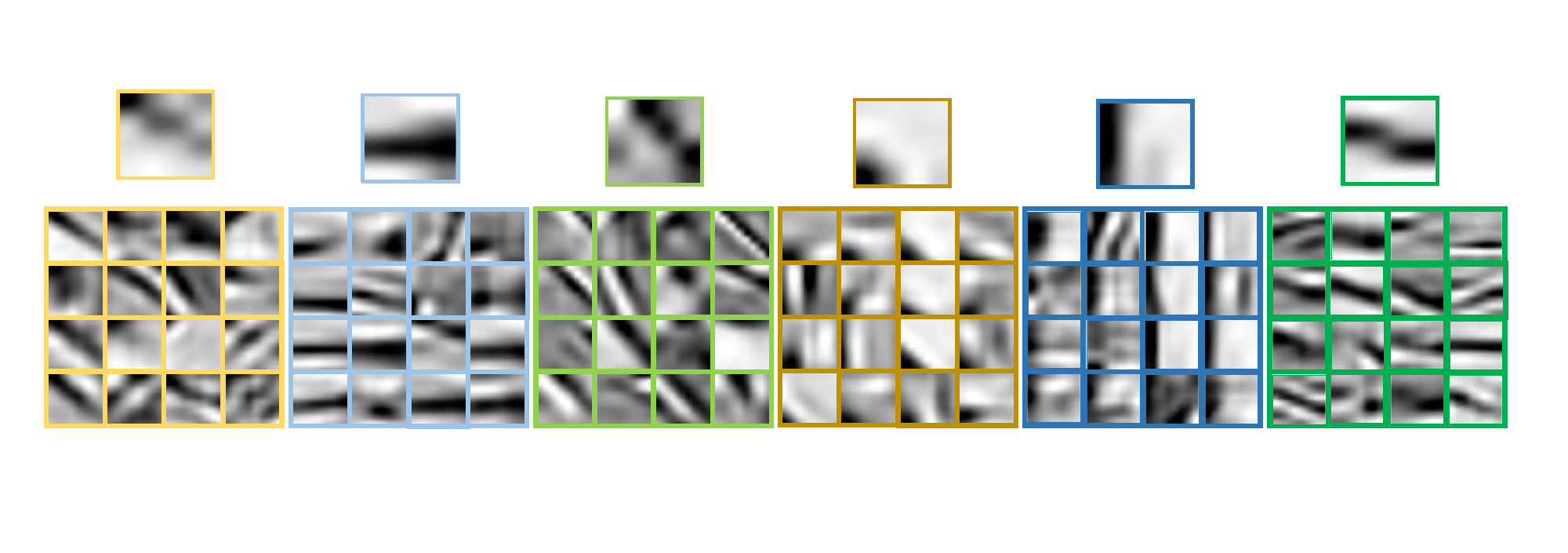}
	\caption{Visualization of select low-resolution atoms and their corresponding atoms in the high resolution dictionary; top -- low resolution atoms, scaled up to show features clearly; bottom -- corresponding high resolution atoms per each low resolution atom. By restricting the support set of higher resolution approximation, our method learns child atoms similar to parent atom.}
	\label{fig:atom_compare}
\end{figure*}

\begin{figure*}[!ttt]
	\centering
	\begin{subfigure}[t]{0.234\textwidth}
		\centering
		\includegraphics[width=\textwidth]{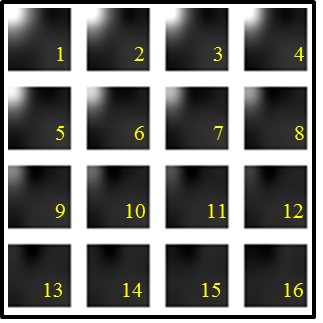}
		\caption{Parent atom (upscaled)}
	\end{subfigure}
	\begin{subfigure}[t]{0.76\textwidth}
		\centering
		\includegraphics[width=0.31\textwidth]{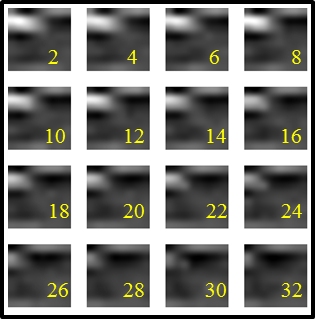}
		\includegraphics[width=0.31\textwidth]{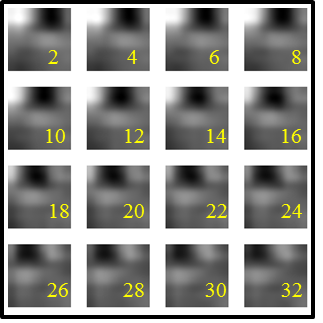}
		\includegraphics[width=0.31\textwidth]{figures/video/atom/1/atom_l0_2.png}
		\caption{Child atoms shown with alternate frames.}
	\end{subfigure}
	\caption{Visualization of frames of a $8\times8\times32$ parent and select children atoms. Notice that the child atoms have similar motion pattern to the parent atom, with added spatial details.}
	\label{fig:video_atoms_high_res}
\end{figure*}
\begin{figure*}[!thh]
	\centering
	\includegraphics[width=0.12\textwidth]{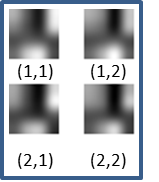}
	\includegraphics[width=0.21\textwidth]{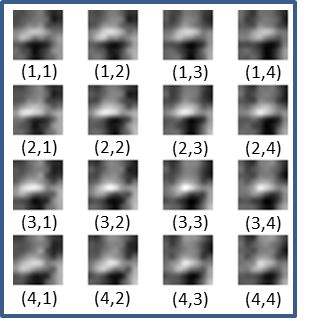}
	\includegraphics[width=0.21\textwidth]{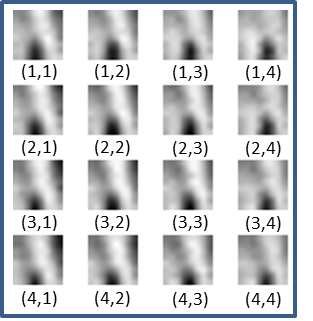}
	\includegraphics[width=0.21\textwidth]{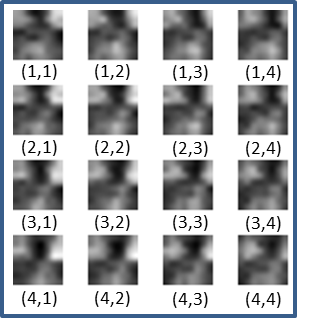}
	\includegraphics[width=0.21\textwidth]{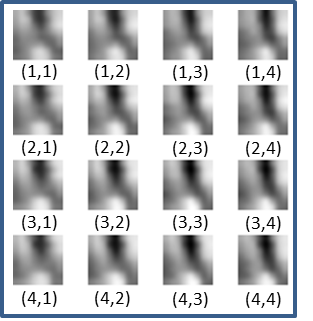}
	\caption{Visualization of a $4\times4\times2\times2$ parent atom (left) and frames of select $8\times8\times4\times4$ high resolution child atoms. The low resolution atoms have been scaled up by a factor of 2 to show features clearly. The various sub-aperture views in child atoms are similar to the parent atoms but with added spatial details.}%, which confirms the ubiquity of our proposed signal model for visual signals.}
	\label{fig:lf_atoms_high_res}
\end{figure*}

\subsection{Initialization of dictionary}
Since the dictionary learning objective as well as the multi-scale dictionary learning objective are non-convex, the solution obtained depends on the initialization. Elad et al. \cite{elad2006image} proposed certain initialization and update heuristics which ensured good results. Along the same lines, we propose the following heuristics:

\begin{enumerate}
	\item The lowest resolution dictionary may be initialized as proposed in \cite{elad2006image}. In our experiments, we initialized by picking $T_\low$ training patches randomly.
	\item For initializing higher resolution dictionaries, we use low resolution dictionary information. Let $D_\low$ be output of the first step of multi-scale dictionary training. Let $\Omega_k =\{i: W\bfx_j \approx \alpha_{j,k} \bfd_{\low, k} +r_j \quad \forall j=1,2,...,n\}$. Then, $\bfd_{\high, j}, \forall j \in f(k)$ is randomly initialized from the sub training samples, $X_{|\Omega_k}$.
	\item An unused atom from the lowest resolution dictionary may be replaced by the least represented training sample, as proposed in \cite{elad2006image}.
	\item Let $\Omega_k$ be as defined above. Then an unused atom in a higher resolution dictionary may be replaced from the least represented training sampling in $X_{\Omega_k}$.
\end{enumerate}

Figures \ref{fig:atom_compare}, \ref{fig:video_atoms_high_res}, and \ref{fig:lf_atoms_high_res} show examples of the learnt low-resolution atoms and the corresponding high-resolution atoms for images, videos and light fields. Observe that constraining the sparse support of the high-resolution approximation alone learns patches which are very similar in appearance to the low-resolution patches, which supports our proposed signal model.

%----------- EXPERIMENTAL RESULTS ------------ %
\section{Experimental results} \label{sec:results}

% Experiments section.
\subsection{Simulation details}
\label{subsec:sim}
To validate our signal model, we show that our signal model performs as good as a large dictionary with runtimes compared to that of a small dictionary. We trained dictionaries over various classes of visual signals to emphasize the ubiquity of our signal model. Comparisons were made against a small dictionary with (1) $N_\low$ atoms, (2) a large dictionary with $N_\high$ atoms, (3) our proposed multiscale dictionary with $N_\low$ low resolution and $N_\high$ high resolution atoms, and (4) a $N_\high$ multi-level dictionary with $N_\low$ levels, as proposed by Jayaraman et al. \cite{thiagarajan2013learning}. We quantify the approximation accuracy using recovered SNR that is defined as follows: given a signal $\bfx$ and its estimate $\widehat{\bfx}$, $\textrm{SNR } = 20 \log_{10}(\| \bfx \| / \| \bfx - \widehat{\bfx} \|)$.

\begin{figure}[ttt]
\centering
\includegraphics[width=0.24\textwidth]{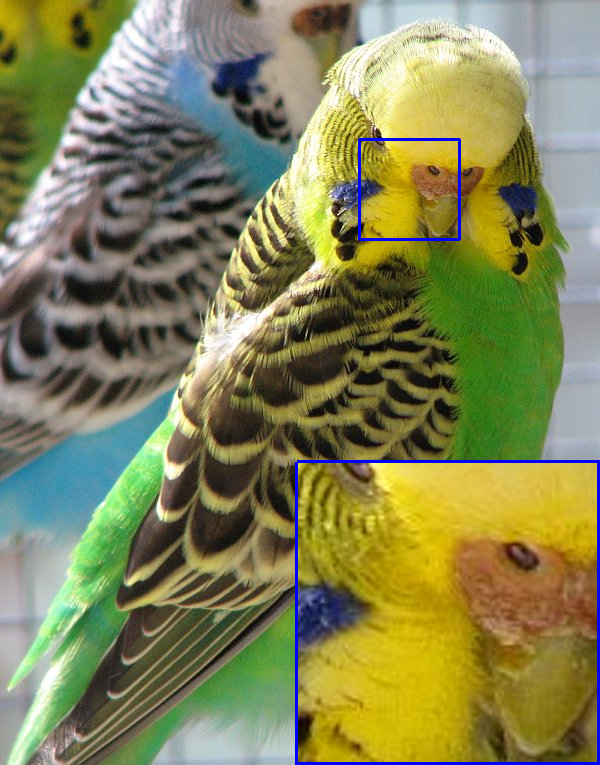}
\includegraphics[width=0.24\textwidth]{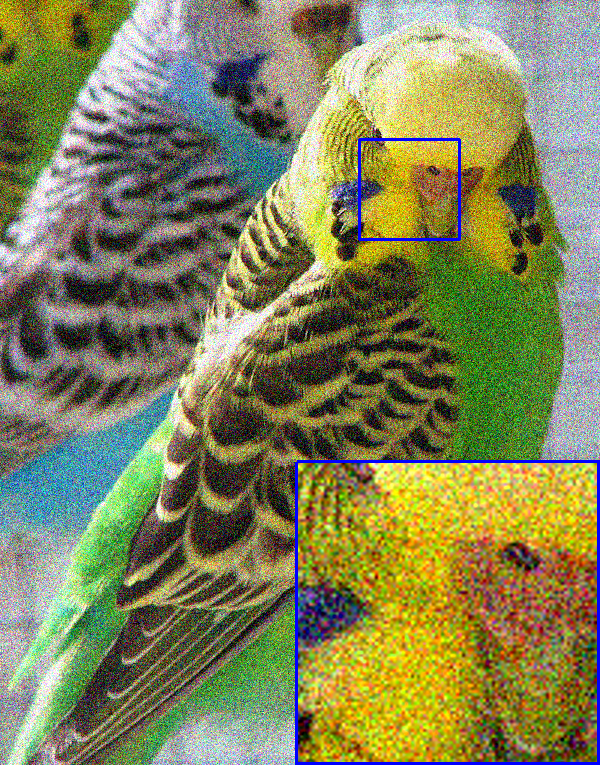}	
\vspace{-0.1em}

\includegraphics[width=0.24\textwidth]{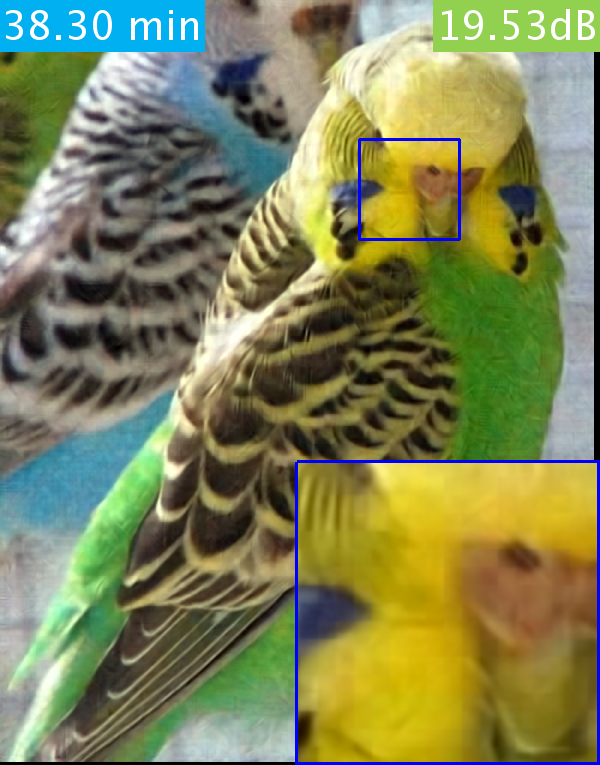}
\includegraphics[width=0.24\textwidth]{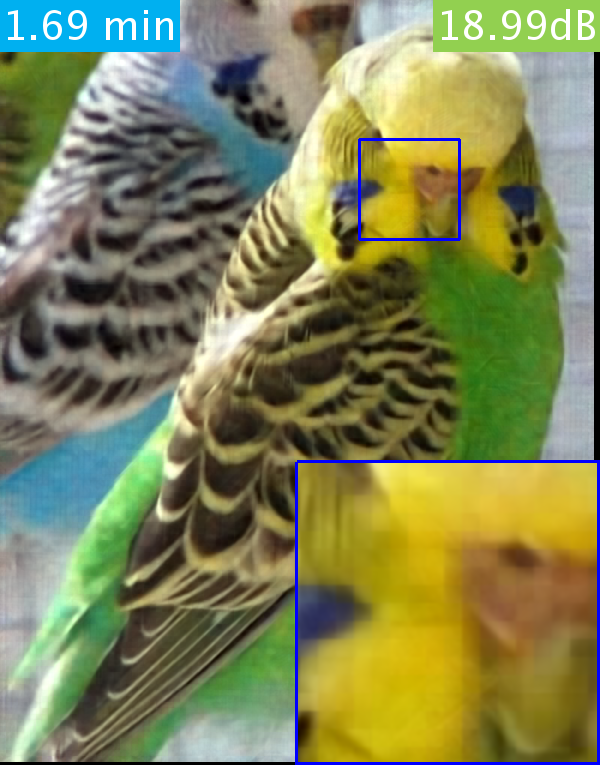}
\caption{Visualization of results for denoising the ``Budgerier" image. Clockwise from top left, original image, noisy image with SNR of $10\textrm{dB}$, recovered image using proposed method, and  recovered image using K-SVD learned dictionary. We obtain a speedup of  $22 \times$ with less than 1dB reduction in accuracy.}
\label{fig:img_visual}	
\end{figure}
\begin{table*}[!ttt]
	\includegraphics[width=\textwidth]{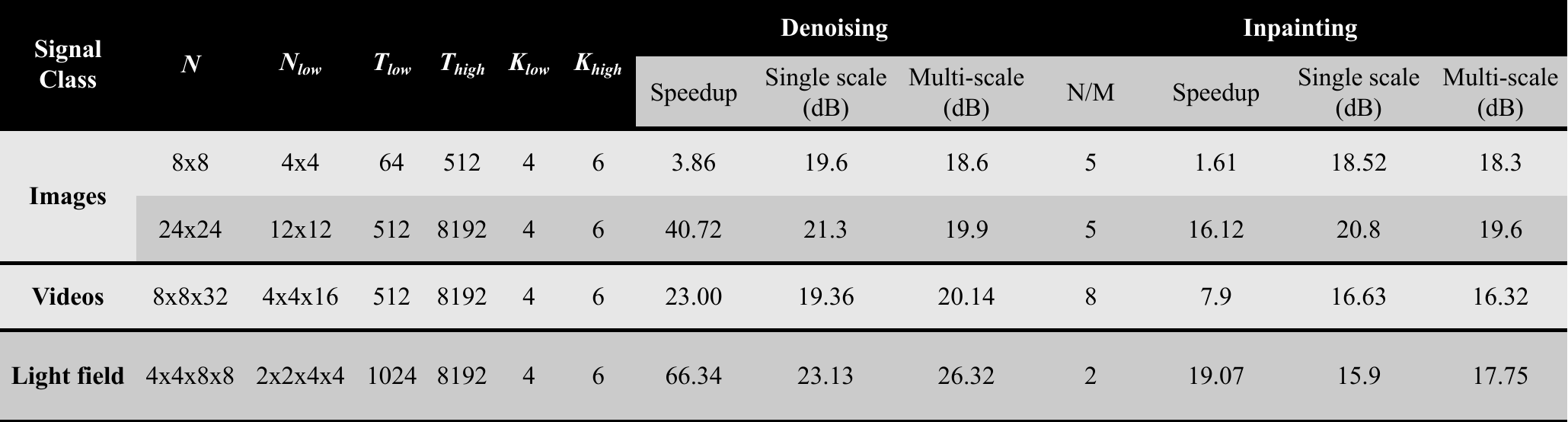}
	\caption{Table with speed up and accuracy for denoising and CS with various class of signal. Denoising was performed with 15dB SNR conditions. $N/M$ represents the number of unknowns per each known variable.}
	\label{table:mix}
\end{table*}

\begin{figure}[!ttt]
	\centering
	\includegraphics[width=0.23\textwidth]{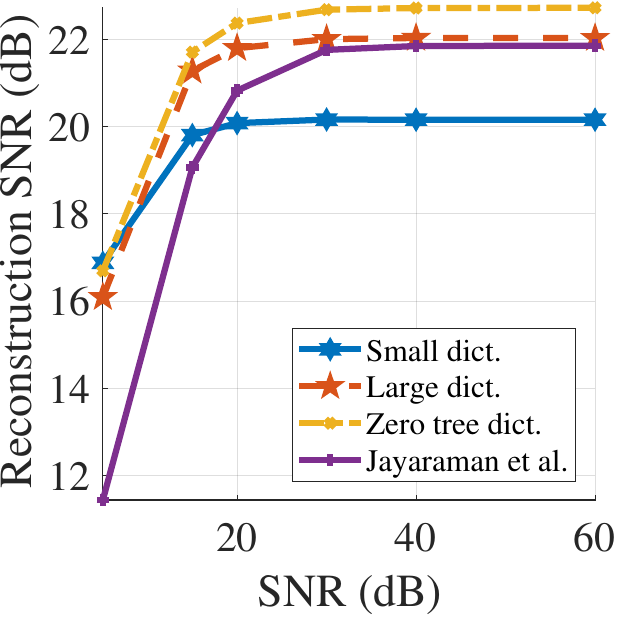}
	\includegraphics[width=0.23\textwidth]{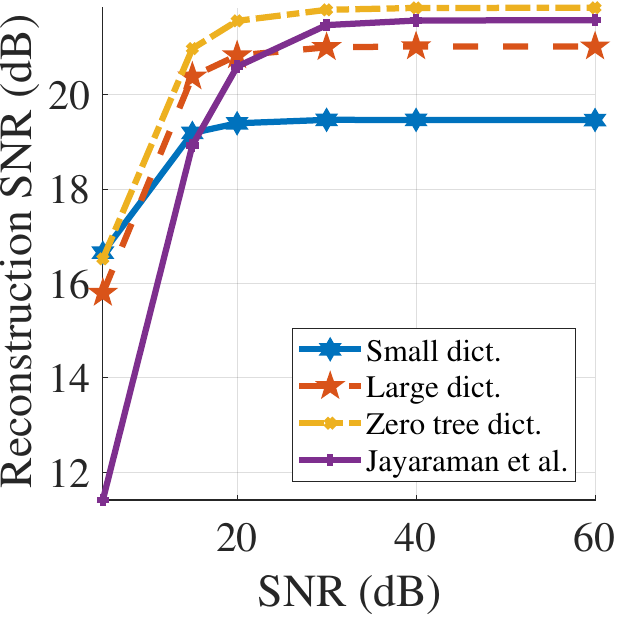}
	\includegraphics[width=0.23\textwidth]{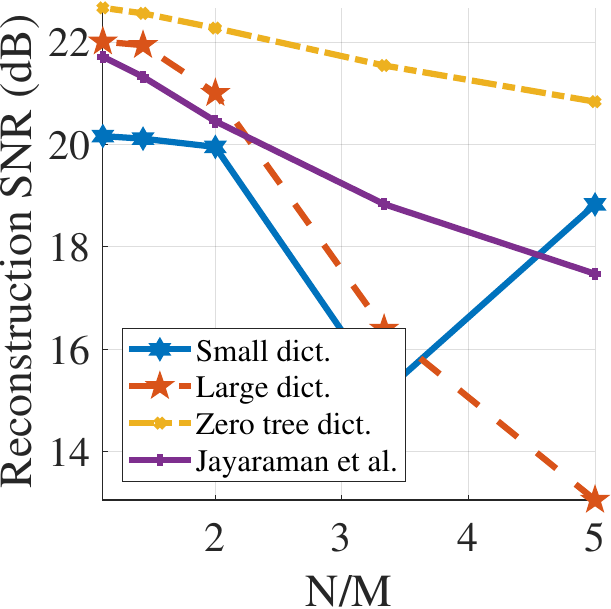}
	\includegraphics[width=0.23\textwidth]{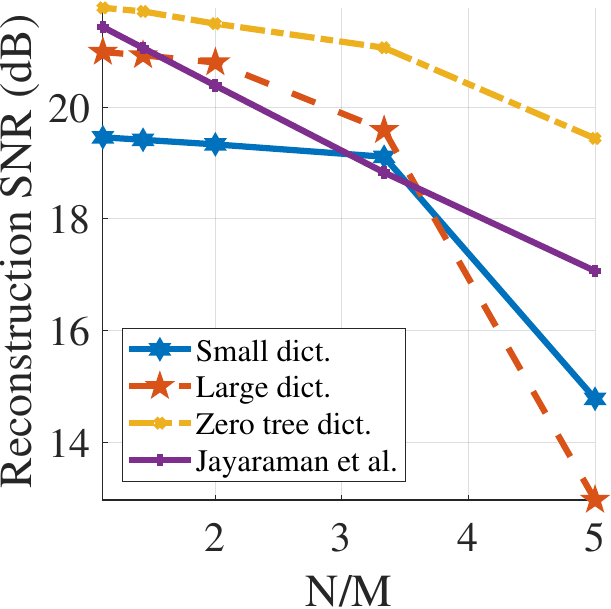}
	\caption{Performance on (top-row) denoising with additive Gaussian noise and (bottom-rown) inpainting with randomly removed pixels for two images --- (left column) peppers and (right) budgerier. Here, $N/M$ represents the number of unknowns per each known. See Section \ref{subsec:sim} for details about each dictionary. The proposed method provides better reconstruction accuracy over competing methods.}
	\label{fig:im_plots}
\end{figure}

\subsection{Images}
We trained dictionaries with $N_\low = 512$ and $N_\high = 8192$ on $24\times24$ image patches and downscaled patches of dimension $12\times12$. 

Figure \ref{fig:bayer} shows demosaicing of the Bayer pattern using a large single scale dictionary and our proposed method.
We trained an $8192$ atom high resolution dictionary on $24\times24$ Kodak True color RGB images\cite{kodak} and $512$ atom low resolution dictionary on the patches downscaled to $12\times12$.
We compare this against $8192$ atom single scale dictionary. It took $16$ minutes for the single scale with an approximation accuracy of 18.45dB, whereas only $1.5$ minutes with an approximation accuracy of 18.43dB for the two scale dictionary. 

Figure \ref{fig:img_visual} shows image denoising at an SNR of $10\textrm{dB}$. We perform denoising with the trained RGB dictionaries of $24\times24$ patch and with a patch overlap of $18$ pixels. With hardly any reduction in accuracy, our method performs $22\times$ faster.
Figure \ref{fig:im_plots} compares the performance of various dictionaries for denoising and CS tasks for two representative images. For CS, we retained known pixel values only at a fraction of the locations and recovered the complete image. For both the cases, our dictionary outperforms other methods. Speed ups obtained for denoising and CS is summarized in Table \ref{table:mix}. With 1dB or less loss in accuracy, our method offers significant speed ups for all image processing tasks.

It is worth mentioning that BM3D \cite{dabov2007image,kostadin2007video}, one of the classical image denoising techniques that uses non-local statistics, provides exceptional denoising results; typically, at 15 dB measurement noise BM3D outperforms most sparse optimization-based denoisers --- including the proposed method --- by 9 dB or so. 
However, the run times associated with BM3D are often longer than  our approach. 
Further, it is also worth noting that the proposed idea as well as most dictionary-based representations are designed towards solving  general linear inverse problems that go beyond denoising. 

\begin{figure}[t!]
	\centering
	\includegraphics[width=0.35\textwidth]{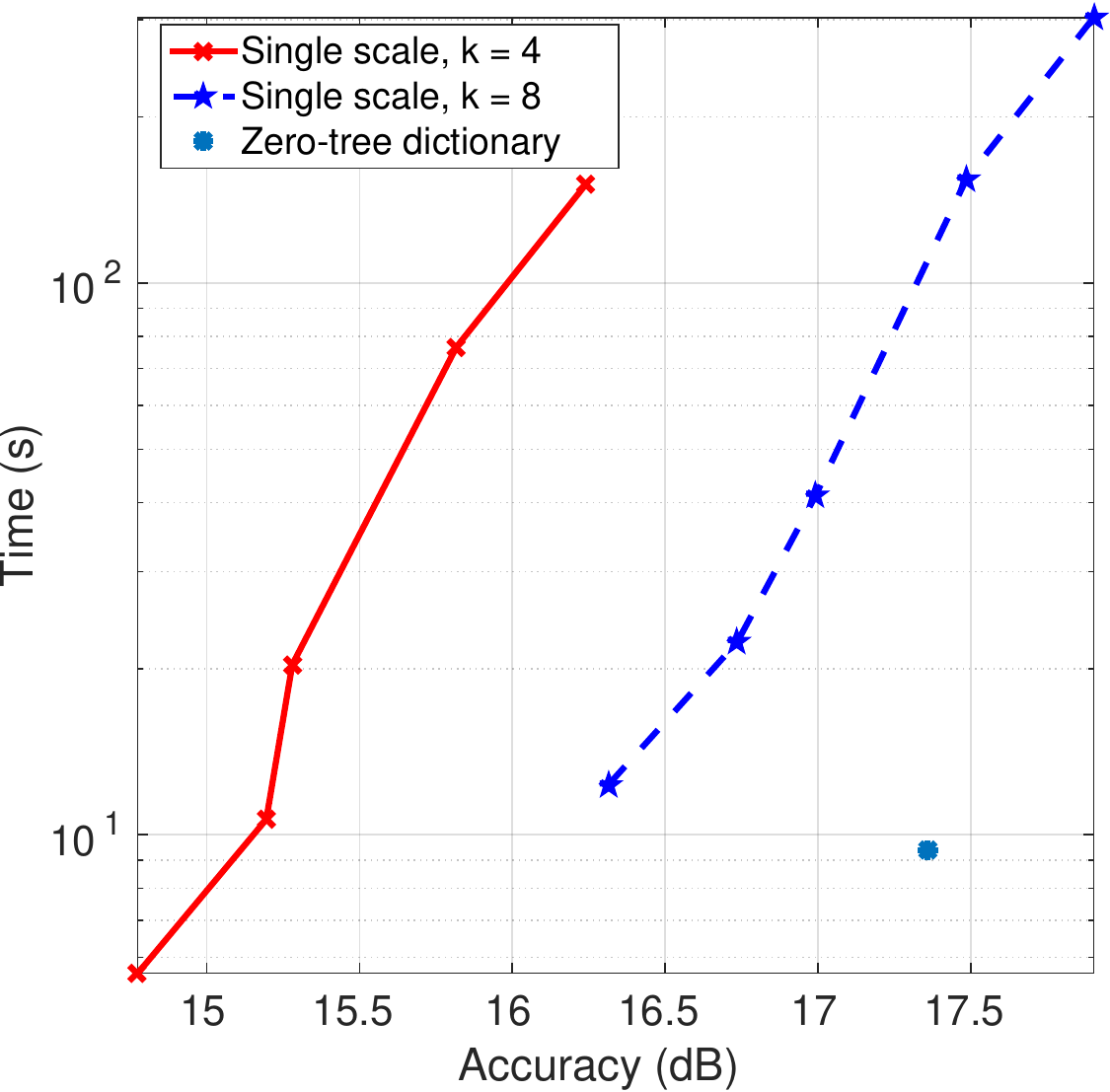}
	\caption{Time vs accuracy  for denoising of videos at $15dB$ input noise, with single scale dictionary and proposed multi-scale dictionary. Dictionaries of sizes 256, 512, 1024, 4096 and 8192 are compared against a zero tree dictionary of 8192 atoms of high resolution and 512 atoms of low resolution, with $K_\low=6$ and $K_\high=8$. At high approximation accuracies, our method outperforms large dictionaries in run-time time.}
	\label{fig:time_accuracy}
\end{figure}
\begin{figure}[!ttt]
	\centering
	\includegraphics[width=0.23\textwidth]{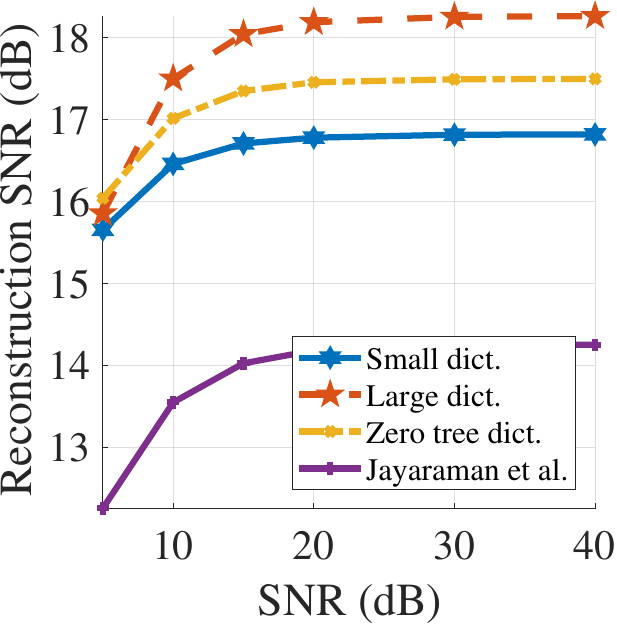}
	\includegraphics[width=0.23\textwidth]{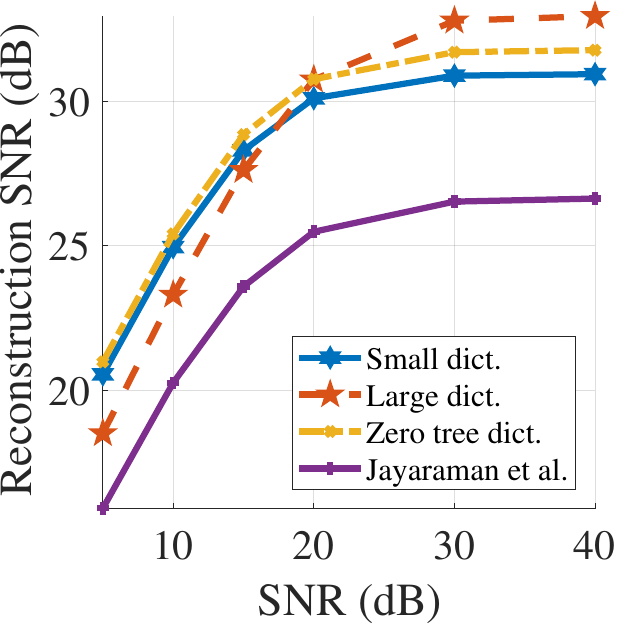}
	\includegraphics[width=0.23\textwidth]{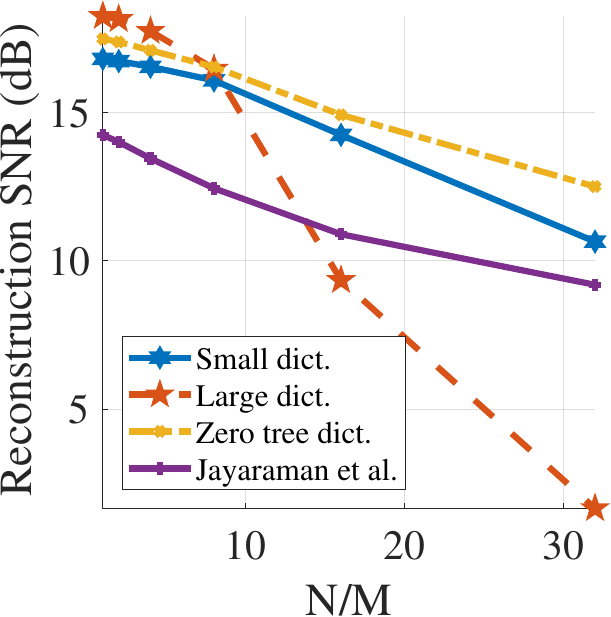}
	\includegraphics[width=0.23\textwidth]{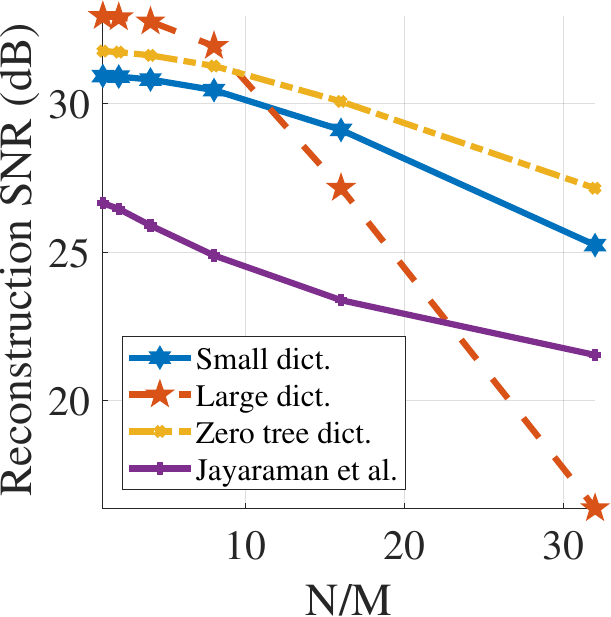}
	\caption{Reconstruction performance on (top-row) denoising with additive Gaussian noise and (bottom) compressive sensing using the imaging architecture of  Hitomi et al.\ \cite{hitomi2011video} for two different videos --- (left) sharpner and (right) egg drop. Here, $N/M$ represents the number of frames reconstructed from each coded image.}
	\label{fig:vid_plots}
\end{figure}

\begin{figure*}[!ttt]
	\centering
	\includegraphics[width=0.3\textwidth]{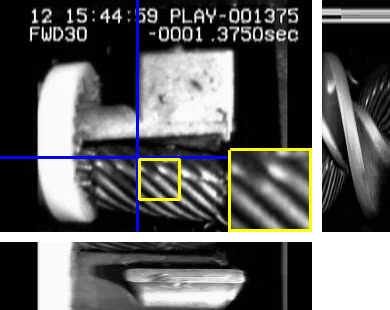}
	\hspace{1em}
	\includegraphics[width=0.3\textwidth]{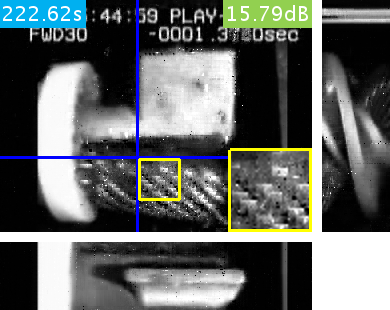}
	\hspace{1em}
	\includegraphics[width=0.3\textwidth]{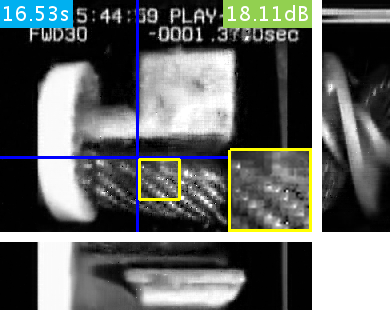}
	
	\vspace{1em}
	\includegraphics[width=0.3\textwidth]{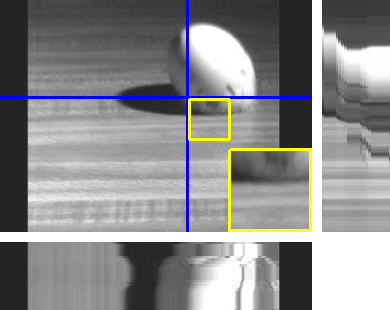}
	\hspace{1em}
	\includegraphics[width=0.3\textwidth]{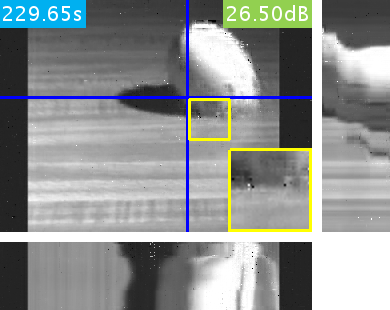}
	\hspace{1em}
	\includegraphics[width=0.3\textwidth]{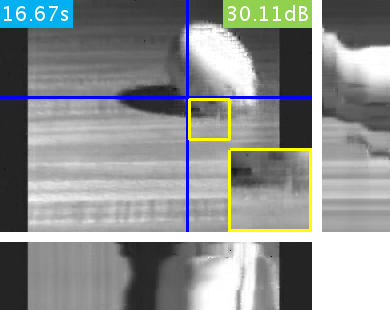}
	\caption{Visualization of reconstructed frames for (top-row) sharpner  and (bottom) egg drop videos; (columns: from left-to-right) ground truth, reconstruction with single scale dictionary; reconstruction with two scale dictionary. The proposed method provides a speedup of $14\times$ over OMP with a modest 2-4 dB improvement in reconstruction SNR.}
	\label{fig:vid_visu}
\end{figure*} 
\subsection{Videos}
\label{para:videos}
We trained dictionaries with $N_\low = 512$ and $N_\high = 8192$ on $8\times8\times32$ video patches and downscaled video patches of dimension $4\times4\times16$.
We show empirically that our signal model outperforms single scale dictionaries in terms of speed and accuracy. Figure \ref{fig:time_accuracy} show the comparison of single scale dictionaries of various size and zero-tree dictionary for denoising of videos. For the same time of approximation, our method gives the highest accuracy. Put it another way, for the same accuracy, our signal model takes the least time.

Figure \ref{fig:vid_plots} shows the performance of various dictionaries for denoising and CS tasks and Figure \ref{fig:vid_visu} show results for CS of videos.
For CS, we combined multiple frames into a coded image, as proposed by Hitomi et al. \cite{hitomi2011video}. Speedup in denoising was $20\times$ while that for CS was between $5\times$ and $15\times$, depending on the number of measurements. Performance for video denoising and CS has been summarized in Table \ref{table:mix}. Speedups obtained for videos is much higher than for images with less than 1dB loss in accuracy. Results are significantly better visually too, as can be seen in Figure \ref{fig:vid_visu},  from the smoother spatial profile compared to reconstruction with single scale dictionary.

\begin{figure}[!ttt]
	\centering
	\includegraphics[width=0.23\textwidth]{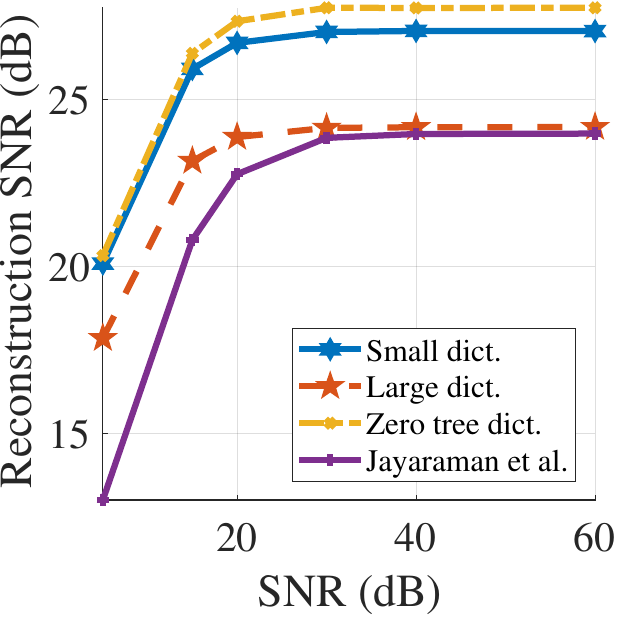}
	\includegraphics[width=0.23\textwidth]{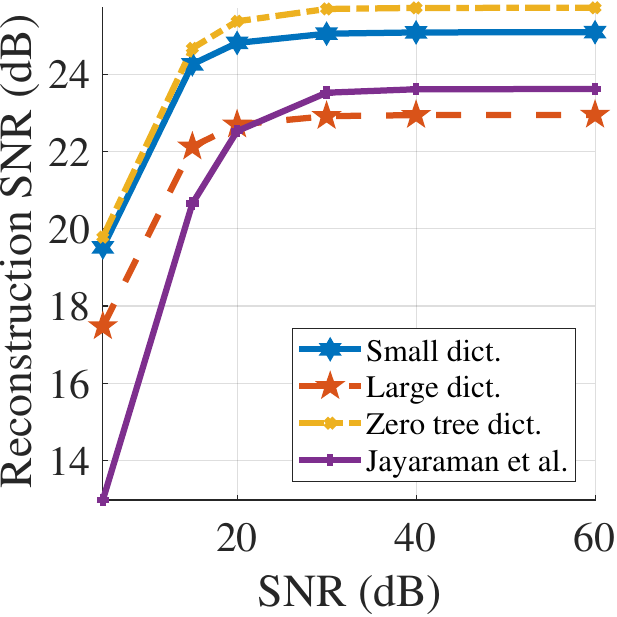}
	\includegraphics[width=0.23\textwidth]{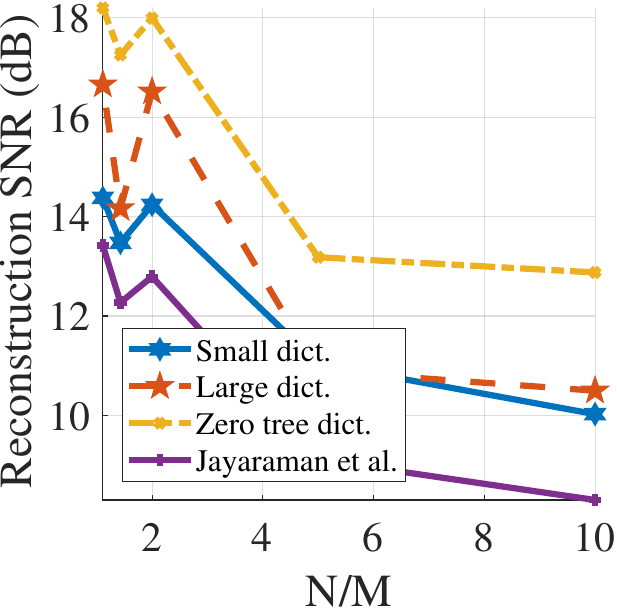}
	\includegraphics[width=0.23\textwidth]{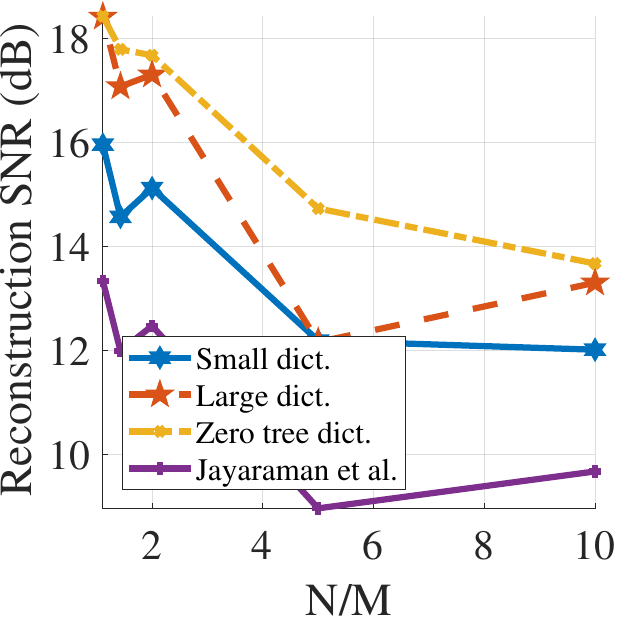}
	\caption{Reconstruction performance on (top row) denoising and (bottom) CS with imaging architecture of Liang et al.\ \cite{liang2008programmable} for  two different light fields -- (left column) Buddha and (right) Dragons. Here, $N/M$ represents the number of sub-aperture views recovered for each coded aperture image.}
	\label{fig:lf_plots}
\end{figure}
\begin{figure*}[!ttt]
	\centering
	\includegraphics[width=0.3\textwidth]{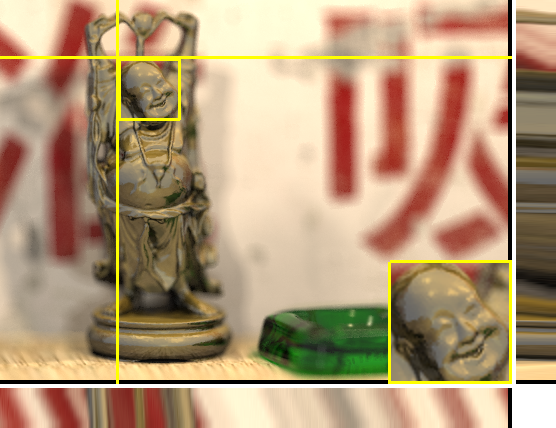}
	\hspace{1em}
	\includegraphics[width=0.3\textwidth]{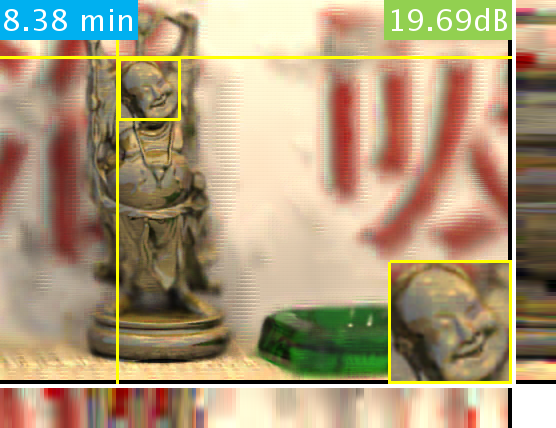}
	\hspace{1em}
	\includegraphics[width=0.3\textwidth]{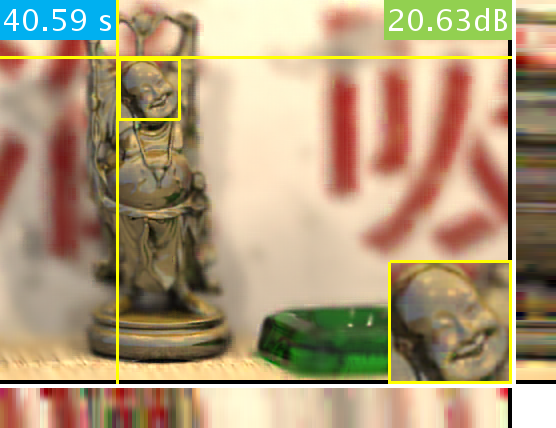}
	
	\vspace{1em}
	\includegraphics[width=0.3\textwidth]{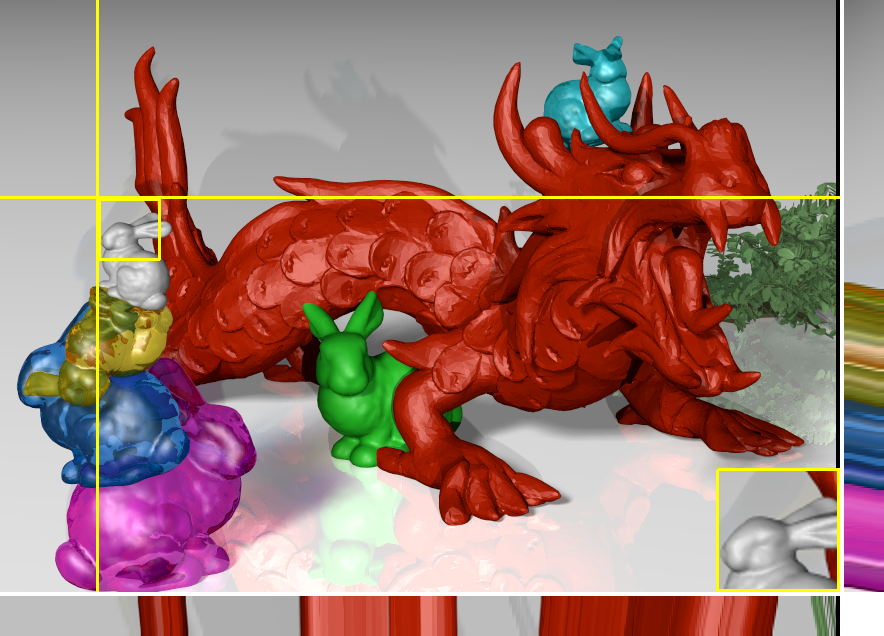}
	\hspace{1em}
	\includegraphics[width=0.3\textwidth]{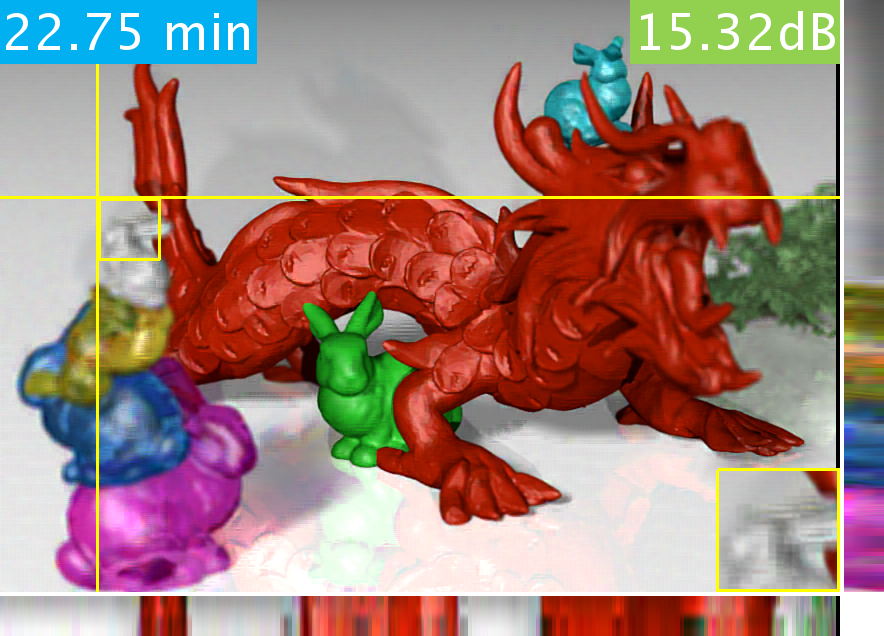}
	\hspace{1em}
	\includegraphics[width=0.3\textwidth]{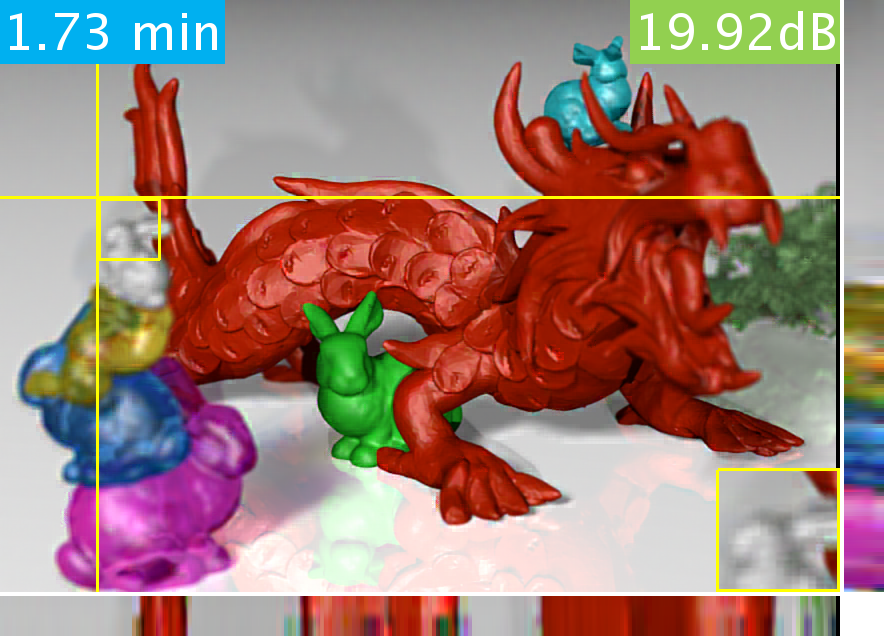}
	\caption{Visualization of reconstructed sub-aperture views for (top-row) Buddha  and (bottom) dragon light fields; (columns: from left-to-right) ground truth, reconstruction with single scale dictionary; reconstruction with two scale dictionary. The epipolar slices have been scaled up to show features clearly. The proposed method enables a  $10\times$ speed up along with an increase of $1-4$ dB in reconstruction SNR.}
	\label{fig:lf_visu}
\end{figure*}
\begin{figure*}[!ttt]
	\begin{minipage}{0.5\textwidth}
	\includegraphics[width=0.46\textwidth]{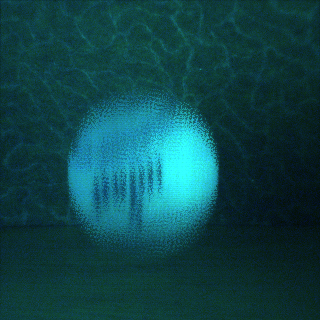}
	\includegraphics[width=0.46\textwidth]{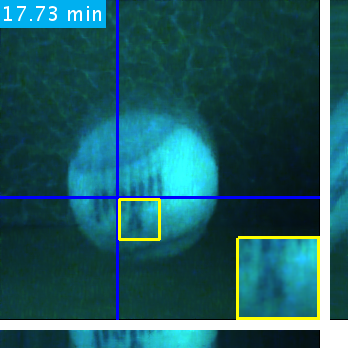}
	\vspace{0.5em}

	\includegraphics[width=0.46\textwidth]{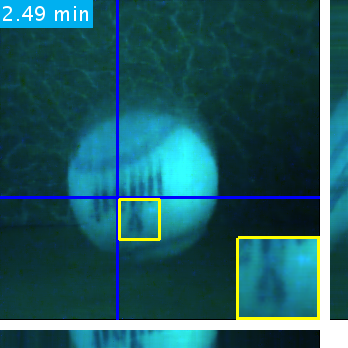}
	\includegraphics[width=0.46\textwidth]{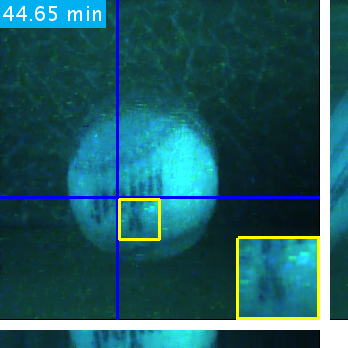}
	\end{minipage}
	\begin{minipage}{0.5\textwidth}
	\includegraphics[width=0.46\textwidth]{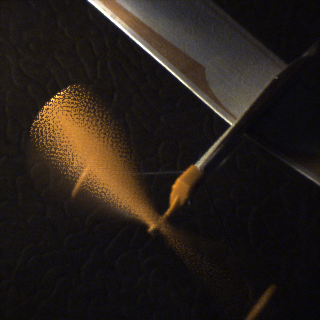}
	\includegraphics[width=0.46\textwidth]{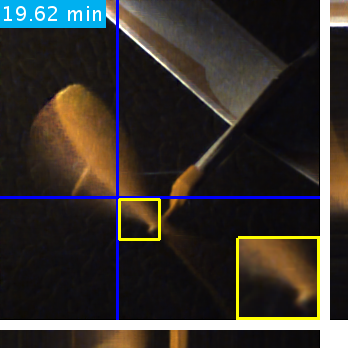}
		\vspace{0.5em}
		
	\includegraphics[width=0.46\textwidth]{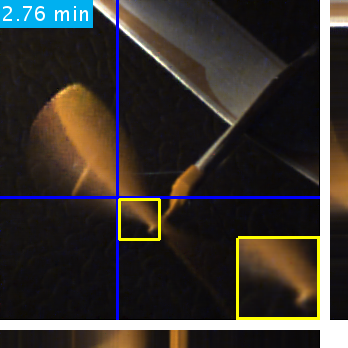}
	\includegraphics[width=0.46\textwidth]{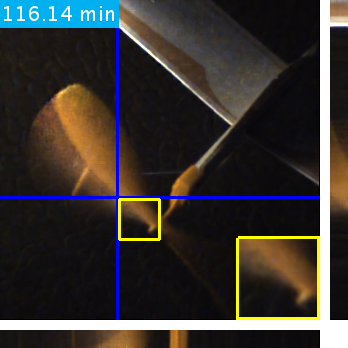}
	\end{minipage}
	\caption{Reconstruction from real data for two scenes from \cite{hitomi2011video}. For each of the scenes, clockwise from top left: coded image, reconstruction with 10,000 atom dictionary, reconstruction with 100,000 atom dictionary and reconstruction with the proposed two-scale dictionary.  The proposed method provides a $6\times$ speed up over the 10,000 atom dictionary at an improved visual quality --- for example, the number ``4" is rendered with fewer artifacts in our reconstruction.}
	\label{fig:vid_ball}
\end{figure*}

\subsection{Light field images}
We trained dictionaries with $N_\low = 512$ and $N_\high = 8192$ on $4\times4\times8\times8$ video patches and downscaled video patches of dimension $2\times2\times4\times4$.
Figure \ref{fig:lf_plots} shows the performance of various dictionaries for denoising and CS tasks and Figure \ref{fig:lf_visu} shows results for CS of light fields.
For CS, we simulated acquisition of images with multiple coded aperture settings, proposed in \cite{liang2008programmable}. Performance metrics for denoising and CS has been summarized in Table \ref{table:mix}. Our method not just offers a $19 - 60\times$ speed up, but there is an increase in reconstruction accuracy. Improved results can be observed visually in the reconstruction results of the Dragons and Buddha datasets in Figure \ref{fig:lf_visu} from compressive measurements.

\subsection{Experiments on real data}
We tested our algorithm on real data collected by Hitomi et al.\ \cite{hitomi2011video}. A $320\times 320\times 18$ video was reconstructed from a $320\times 320$ coded image. We compared reconstruction with our zero-tree dictionary and the dictionaries of 10,000 and 100,000 atoms trained by Hitomi et al. Figure \ref{fig:vid_ball}  show the reconstruction results for a bouncing ball and a toy airplane respectively with the time taken for reconstruction. Notice that the results are visually similar while our method is faster by $6\times$ compared to the stock 10,000 atom dictionary. In case of the bouncing ball, the number ``4" has been better resolved in our result as well. 

\begin{table*}[!hhh]
	\includegraphics[width=\textwidth]{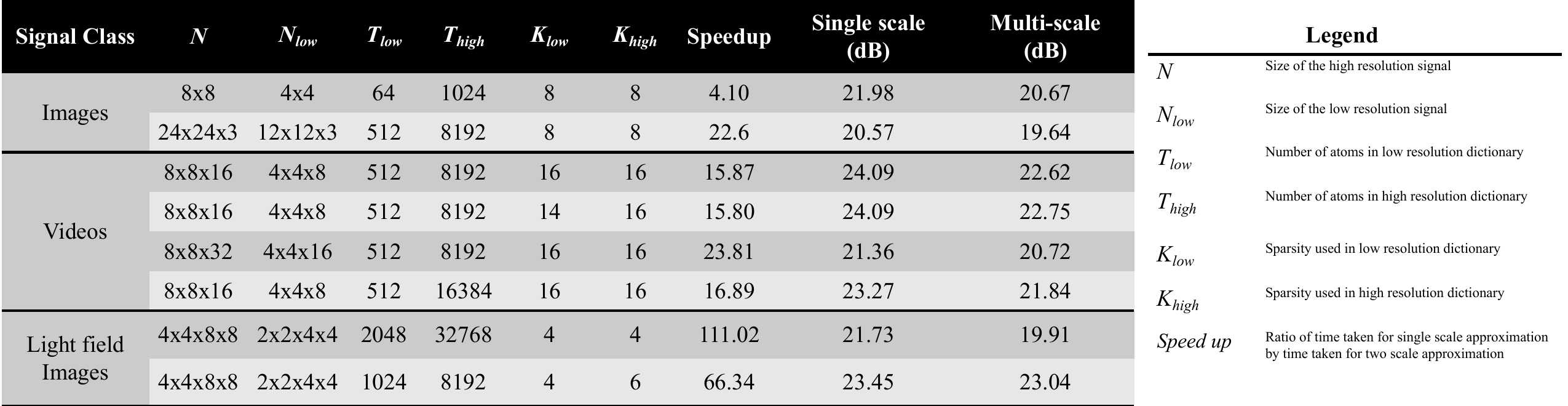}
	\caption{Table with speed up for various dictionary sizes, patch sizes and sparsity. The speed up shown are for solving sparse approximation problems and quantify the ratio of time taken by OMP using a K-SVD learnt dictionary to  zero tree OMP on the proposed model. Also shown are approximation errors on training dataset for both K-SVD and the proposed  algorithm.}
	\label{table:model}
\end{table*} 

\subsection{Summary}
Table \ref{table:model} and Figures \ref{fig:im_plots}, \ref{fig:vid_plots} and \ref{fig:lf_plots} quantify the performance of the proposed signal model and those obtained using K-SVD for a wide range of parameters as well as signals. 
Across the board, we observe that the proposed framework provides accuracies that are as good as those obtained with K-SVD, but with speedups that are  $4-10\times$  for small-sized problems  and  $20-110\times$ for larger problems. The speedups obtained are comparable to results in \cite{ayremlou2014fast} with higher approximation accuracies for our proposed method. 
As a result of speed up of the sparse coding step, we also get significant speed ups during the training phase ($2-40\times$) using modified K-SVD, which makes it feasible to deal with very large problems.

%----------- DISCUSSIONS ---------------------- %
\section{Conclusion and Discussions}
% Conclusion
We presented a signal model that enables cross scale predictability for visual signals.
Our method is particularly appealing because of the simple extension to the existing OMP and K-SVD algorithms while providing significant speed ups at little or no loss in accuracy.
The computational gains provided by our algorithm are especially significant for  problems involving high-dimensional dictionaries with a large number of atoms.
We also believe that the proposed cross scale predictive models can be incorporated with other structural modifications to sparse dictionaries. 
One such example is that of convolutional sparse coding \cite{yang2010supervised,heide2015fast,bristow2013fast}, where each dictionary atom is a convolutional filter and, unlike a patch-based method, the method approximates the sparse coefficients for an  entire image. 
Using cross scale predictive modeling on top would in principle lead to runtime speedups.

\subsection{Limitations}
In order to get higher accuracy of construction, the sparsity levels need to be higher than that for large scale dictionaries with the same number of atoms as the high resolution dictionary. However, this is not a major drawback, as the speedups are still significant in spite of the increased sparsity levels.

Table \ref{table:model} shows that the model accuracy for our proposed signal model is lower than that of large dictionary. This is due to two reasons:
\begin{enumerate}[leftmargin=*]
	\item K-SVD algorithm, being a non-convex optimization framework, is very sensitive to initialization. The initialization proposed in this paper is at best a heuristic. Better results can be obtained with better initialization methods.
	
	\item The dictionary update step for the proposed modified K-SVD algorithm runs independent of the sparse approximation step. A better approach would be to modify the dictionary update step to incorporate the interdependence of sparse coefficients.
\end{enumerate}

\subsection{Connections to  super resolution using dictionaries} Roman et al.\ \cite{zeyde2012single} learn a pair of low resolution and high resolution dictionary using the same sparsity pattern for the two dictionaries. Given a low resolution patch $\bfy_\low$, the sparse approximation problem $\bfy_\low \approx D_\low \bfs$ is first solved and subsequently  the low-resolution image is super resolved as $\bfy=D_\high \bfs$.
In contrast, our method requires the high resolution image as an input and uses the sparse representation of the downscaled image to predict the high resolution sparse representation. 
While the primary aim of \cite{zeyde2012single} is image-based super resolution, our method can accommodate any inverse problem based on sparse approximation. 

\section*{Acknowledgement}
This work has been supported in part by Intel ISRA on Compressive Sensing as well as the NSF CAREER grant CCF-1652569. The authors thank Dengyu Liu for sharing the real experiments data in \cite{hitomi2011video}. The authors also thank Tsung-Han Lin, Gokce Keskin, Chengjie Gu, and Yanjing Li for valuable discussions.

% References
\bibliographystyle{IEEEtran}
\bibliography{refs}

\end{document}